\documentclass[10pt,twocolumn,letterpaper]{article}

\usepackage{iccv}
\usepackage{times}
\usepackage{epsfig}
\usepackage{graphicx}
\usepackage{amsmath}
\usepackage{amssymb}
\usepackage{xspace}
\usepackage{url}
\usepackage{tabularx}
\usepackage{epstopdf}
\usepackage{multirow}
\usepackage{bm}
\usepackage{graphicx}
\usepackage{graphics}
\usepackage[applemac]{inputenc}
\usepackage{cite}
\usepackage{mdwtab}
\usepackage{subfigure}
\usepackage{color}
\usepackage{booktabs}
\usepackage{bbding}

\usepackage{pifont}
\newcommand{\cmark}{\ding{51}}%
\newcommand{\xmark}{\ding{55}}%

\usepackage{array}
\usepackage[percent]{overpic}

\usepackage[titletoc,title]{appendix}

\usepackage[pagebackref=true,breaklinks=true,letterpaper=true,colorlinks,bookmarks=false]{hyperref}

\iccvfinalcopy 

\def\iccvPaperID{609} 

\newcolumntype{C}[1]{>{\centering}m{#1}}

\makeatletter
\DeclareRobustCommand\onedot{\futurelet\@let@token\@onedot}
\def\@onedot{\ifx\@let@token.\else.\null\fi\xspace}

\def\eg{\emph{e.g}\onedot} 
\def\ie{\emph{i.e}\onedot} 
\def\cf{\emph{cf}\onedot} 
\def\etc{\emph{etc}\onedot} \def\vs{\emph{vs}\onedot}
\def\wrt{w.r.t\onedot} 

\makeatother

\newcommand{\Fig}{Fig.\xspace}
\newcommand{\Figs}{Figs.\xspace}
\newcommand{\Sec}{Sec.\xspace}
\newcommand{\Tab}{Tab.\xspace}

\newcolumntype{L}[1]{>{\raggedright\let\newline\\\arraybackslash\hspace{0pt}}m{#1}}
\newcolumntype{C}[1]{>{\centering\let\newline\\\arraybackslash\hspace{0pt}}m{#1}}
\newcolumntype{R}[1]{>{\raggedleft\let\newline\\\arraybackslash\hspace{0pt}}m{#1}}

\newcommand{\MOTNEW}{\emph{MOT16}\xspace}
\newcommand{\MOTOLD}{\emph{MOT15}\xspace}

\newcommand{\myparagraph}[1]{\paragraph{#1}}

\graphicspath{{../figures/}}

\definecolor{darkgreen}{rgb}{0,.75,0}
\definecolor{gray40}{gray}{.40}

\begin{document}

\title{Tracking the Trackers: \\An Analysis of the State of the Art in Multiple Object Tracking}

\author{Laura Leal-Taix\'{e}$^{1,}$\thanks{denotes equal contribution.}  \enspace  Anton Milan$^{2,*}$  \enspace Konrad Schindler$^3$ \enspace Daniel Cremers$^1$ \enspace Ian Reid$^2$ \enspace Stefan Roth$^4$\\
	\small 	$^1$Technical University Munich, Germany \qquad
	\small 	$^2$University of Adelaide, Australia \\
	\small 	$^3$Photogrammetry and Remote Sensing, ETH Z\"urich, Switzerland \qquad
	\small 	$^4$TU Darmstadt, Germany
}



\maketitle 

\newcommand{\thumbwidth}{0.135\linewidth}
\newcommand{\thumbheight}{0.08\linewidth}

\begin{abstract}

Standardized benchmarks are crucial for the majority of computer vision applications.
Although leaderboards and ranking tables should not be over-claimed, benchmarks often
provide the most objective measure of performance and are therefore important guides 
for research.
We present a benchmark for Multiple Object Tracking launched in the late 2014, 
with the goal of creating a framework
for the standardized evaluation of multiple object tracking methods. 
This paper collects the two releases of the benchmark made so far, and provides an in-depth analysis
of almost 50 state-of-the-art trackers that were tested on over 11000 frames.
We show the current trends and weaknesses of multiple people tracking methods, and provide pointers of what researchers should be focusing on to push the field forward.
\end{abstract}

\vspace{-0.3cm}
\section{Introduction}
\label{sec:introduction}

Evaluating and comparing multi-target tracking methods is not trivial 
for numerous reasons (\cf\eg\cite{Milan:2013:CVPRWS}). 
First, unlike for other tasks, such as image restoration, the ground 
truth, \ie~the perfect solution one aims to achieve, is
difficult to define clearly. Partially visible, occluded, or cropped targets, 
reflections in mirrors or windows, and objects that very closely resemble 
targets all impose intrinsic ambiguities, such that even humans may 
not agree on one particular ideal solution. Second, a number of different
evaluation metrics with free parameters and ambiguous definitions often 
lead to conflicting quantitative results across the literature. Finally, the 
lack of pre-defined test and training data makes it difficult to compare
different methods in a fair way.

Even though multi-target tracking is a crucial problem in scene understanding, 
until recently it still lacked large-scale benchmarks to provide a fair comparison between 
tracking methods. 
Typically, methods are tuned to each individual sequence, reaching over 90\% accuracy on well-known sequences like 
PETS \cite{Ferryman:2010:PETS}. Nonetheless, the real challenge for a tracking system is to be able to perform well on 
a variety of sequences with different crowdedness levels, camera motion or illumination, 
ideally with a fixed set of parameters for all sequences.

\renewcommand{\thumbwidth}{0.235\linewidth}
\renewcommand{\thumbheight}{0.15\linewidth}
\begin{figure}[t]
\centering
  \includegraphics[width=\thumbwidth,height=\thumbheight]{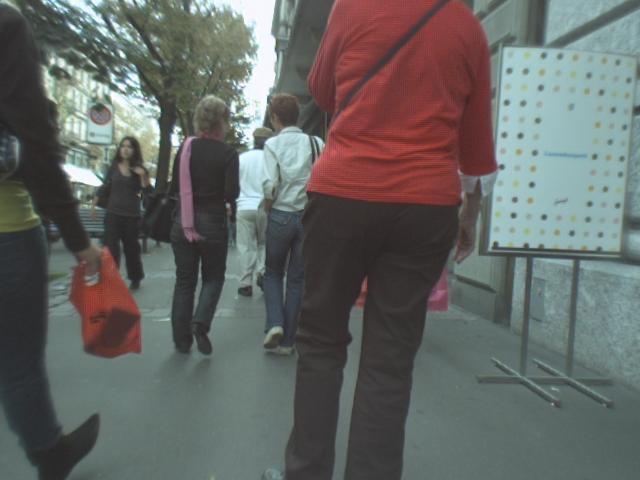}
  \includegraphics[width=\thumbwidth,height=\thumbheight]{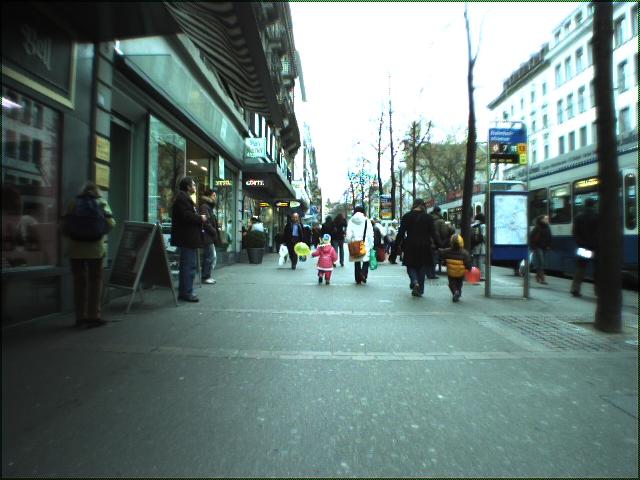}
  \includegraphics[width=\thumbwidth,height=\thumbheight]{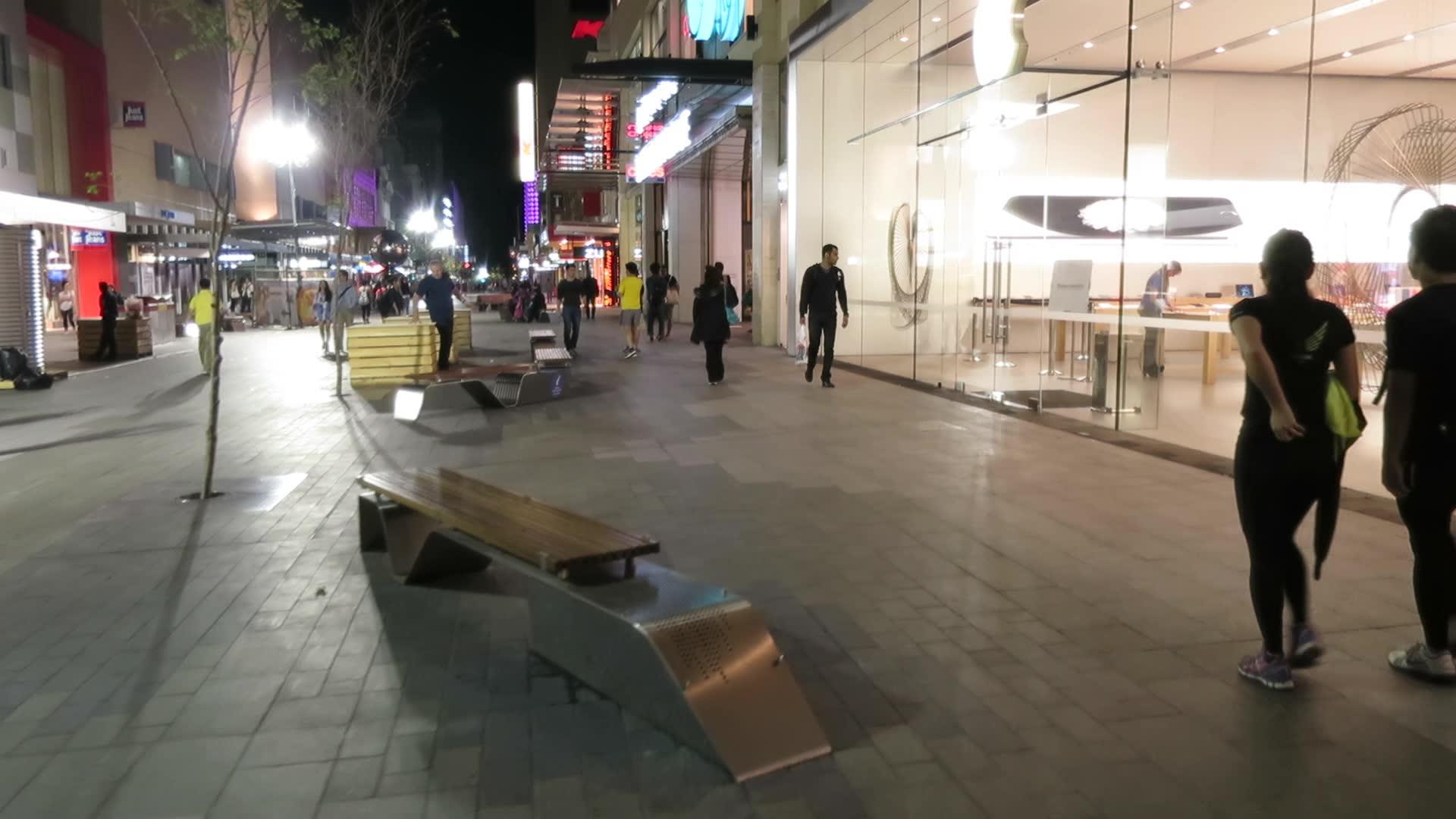}
  \includegraphics[width=\thumbwidth,height=\thumbheight]{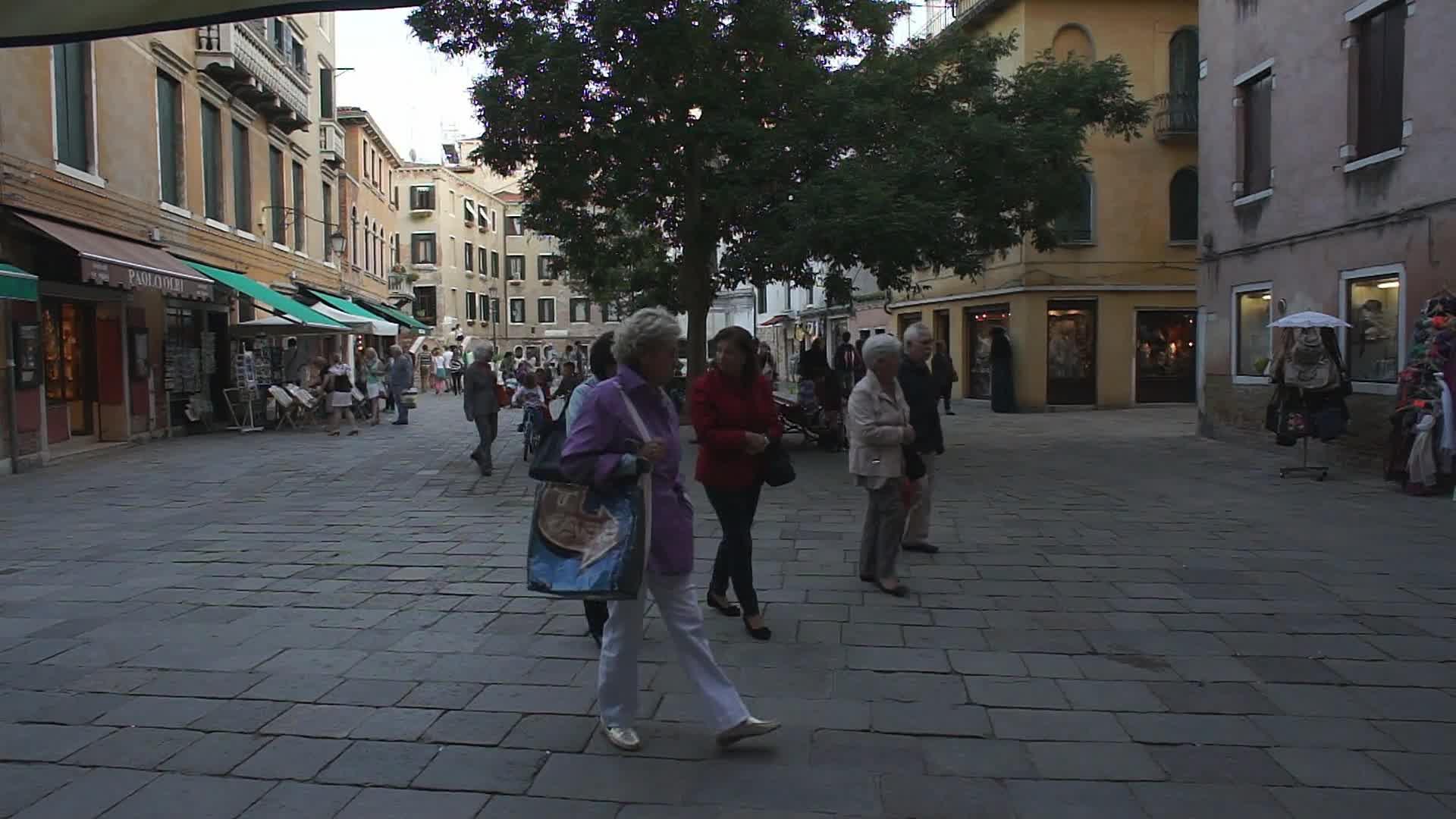}\\[.5em]
 \includegraphics[width=\thumbwidth,height=\thumbheight]{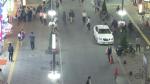}
 \includegraphics[width=\thumbwidth,height=\thumbheight]{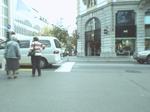}
 \includegraphics[width=\thumbwidth,height=\thumbheight]{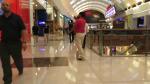}
 \includegraphics[width=\thumbwidth,height=\thumbheight]{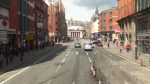}
  \caption{
  Examples frames of the training set for the \MOTOLD release (top) and for the \MOTNEW release (bottom).}
 \label{fig:thumbnails}
 \vspace{-1.5em}
\end{figure}

In order to address this issue, we released a Multiple Object Tracking benchmark \emph{(MOT}) in 2014.
The first release of the dataset, which we refer to as \MOTOLD in this paper, consists of a total of 22 sequences with 101345 annotated pedestrian bounding boxes. 
%
The second release of the dataset, referred to as \MOTNEW, contains a set of 14 distinctively more crowded scenes, with 292733 annotated pedestrian boxes. 
A strict annotation protocol was followed to obtain accurate bounding boxes not only of pedestrians, but of a total of 12 classes of objects, including vehicles, occluders, sitting people, or motorbikes.


Since 2014, hundreds of participants
have submitted their results on a standardized evaluation server, allowing for a fair comparison.
In this work, we analyze 32 published trackers that have been evaluated on \MOTOLD and 16 trackers on \MOTNEW.\footnote{In this paper, we only consider published non-anonymous trackers that were public before 1st of March 2017. We omit LDCT~\cite{SoleraICCV2015} since it is only tested on a subset of the entire benchmark.}
Having results on such a large number of sequences allows us to perform a thorough analysis on trends in tracking, currently best performing methods, and special failure cases. 
We aim to shed some light onto what are current weaknesses in tracking methods and therefore what to focus on for the near future in order to further improve tracking.
Furthermore, we investigate the possibility of creating a `super tracker' by selecting the best tracker for a batch of frames.
Finally, we analyze the evaluation metrics themselves, by conducting an experiment with human observers and analyzing the correlation of various metrics. 

%


This paper has four main goals:
\emph{(i)} To introduce a standardized benchmark for fair evaluation of multi-target tracking methods, along with its two data releases, \MOTOLD and \MOTNEW;
\emph{(ii)} to analyze the performance of 32 state-of-the-art trackers on \MOTOLD and 16 trackers on \MOTNEW;
\emph{(iii)} to analyze common evaluation metrics using an experiment with human evaluators;
  \emph{(iv)} to find the main weaknesses of current trackers and provide pointers to what the community should focus on to advance the field of multi-target tracking.
The main insights gained from the analysis are:
\emph{(1)} Tracker performance is mainly influenced by the affinity metrics used, with deeply learned models giving the most encouraging results;
\emph{(2)} the expected performance of different approaches is highly correlated across videos, \ie most methods perform similarly well or poorly on the same video sequence or fragment;
\emph{(3)} despite some limitations, MOTA remains the most representative measure that coincides to the highest degree with human visual assessment.

%
%
%
%
%
%
%

\section{Related Work}
\label{sec:related-work}
\paragraph{Benchmarks and challenges.}

In the recent past, the computer vision community has developed
centralized benchmarks for numerous tasks including object detection 
\cite{Everingham:2012:VOC}, pedestrian detection 
\cite{Dollar:2009:CVPR}, human pose estimation~\cite{Andriluka:2014:CVPR},
3D reconstruction \cite{Seitz:2006:CVPR}, 
optical flow \cite{Baker:2011:IJCV,Geiger:2012:CVPR,Butler:ECCV:2012}, visual odometry 
\cite{Geiger:2012:CVPR}, single-object short-term tracking 
\cite{VOC2014}, stereo estimation \cite{Scharstein:2002:IJCV, 
Geiger:2012:CVPR}, and video object segmentation \cite{perazzicvpr2016}, among others. 
Despite potential pitfalls of such benchmarks (\eg~\cite{Torralba:2011:ULD}), they 
have proven to be extremely helpful in advancing the state of the art in 
the respective area. For multiple target tracking, in contrast, there
has been very limited work on standardizing quantitative
evaluation.

One of the few exceptions is the well known PETS dataset 
\cite{Ferryman:2010:PETS}, targeted primarily at surveillance 
applications with subtasks like person counting, density estimation, flow analysis, or event recognition.
Even for this
widely used benchmark, we observe that tracking results are commonly
obtained in an inconsistent way: using different subsets of 
the available data, inconsistent model training that is often prone to overfitting, varying evaluation scripts, and different detection inputs.
Results are thus not easily comparable.

\begin{table}[tb]
\centering
\caption{Overview of the characteristics of the data releases including both train and test splits.}
\smallskip
\resizebox{1\linewidth}{!} {
\begin{tabular}{@{}l ccccccc @{}}
\toprule
  Release & \# Seq. & BBs & Persons & Length & Density & Tracks &  HD    \\ \midrule
 \MOTOLD & 22 & 101349 & 101349 & 16:36 & 8.98 & 1221 & 31.8$\%$ \\
 \MOTNEW & 14 & 476532 & 292733 & 07:43 & 26.05 & 1276 & 85.7$\%$    \\
\bottomrule
\end{tabular}
}
\label{tab:datasets}
\vspace{-0.5em}
\end{table}
A well-established and useful way of organizing datasets is through 
standardized challenges, where results are computed in a centralized way, making comparison with any
other method immediately possible.
There are several datasets organized in this fashion: Labeled Faces 
in the Wild \cite{huangtech2007} for unconstrained face recognition,  
PASCAL VOC \cite{Everingham:2012:VOC} for object detection, scene classification and semantic segmentation, the DAVIS challenge \cite{perazzicvpr2016} for video object segmentation, or the ImageNet 
large scale visual recognition challenge \cite{Russakovsky:2014:ImageNET}.
The KITTI benchmark \cite{Geiger:2012:CVPR} contains 
challenges in autonomous driving, including stereo/flow, odometry, 
road and lane estimation, object detection and orientation estimation, as 
well as tracking. Some of the sequences include crowded pedestrian 
crossings, making the dataset quite challenging. However, this dataset is heavily biased towards autonomous driving applications, only showing street scenes captured from a fixed camera of a driving or standing car.
Cityscapes~\cite{Cordts:2016:Cityscapes} is a more recent benchmark, targeting semantic and instance-level segmentation in high-definition videos, but like KITTI, is targeted for urban environments only.

With our benchmark, we want to provide a highly diverse and more challenging set of video sequences to push the limits of current multi-target tracking approaches. As can be seen in \Fig~\ref{fig:thumbnails}, our dataset includes videos from both static and moving cameras, low and high image resolution, varying weather conditions and times of the day, viewpoints, pedestrian scale, density, and more.






\myparagraph{Analysis of the state of the art.} 
Existing benchmarks help to push research forward and find out weaknesses of current methods for several tasks. \cite{Dollar:2012:PAMI} provides a thorough analysis of the state of the art in pedestrian detection on the Caltech Pedestrians dataset~\cite{Dollar:2009:CVPR}, focusing on the results of several detectors and analyzing their performance in detail.
Depth estimation and optical flow evaluation were first standardized with the Middlebury dataset and the results of several methods were compared in~\cite{Baker:2011:IJCV}. 
The results from 5 years of the PASCAL VOC challenge are summarized and analyzed thoroughly in\cite{Everingham:2015:IJCV}. A recent study accompanying the ImageNet recognition challenge~\cite{Russakovsky:2014:ImageNET} also analyzes several methods that participated in that challenge, showing what tasks can still be improved and what are the main problems of current methods.
The Visual Object Tracking (VOT) Challenge~\cite{Kristan:2016:VOT} for single object tracking, releases a yearly report with an analysis of all results presented in that year, showing the steady improvement in visual tracking. 
Similar to all these works, our goal is to establish a meaningful benchmark and to conduct a thorough analysis of the state of the art in multiple object tracking, providing valuable insights to the community.

%

%

\section{The Multiple Object Tracking Benchmark}
\label{sec:datasets}

One of the key aspects of any benchmark is data collection. The goal of our benchmark is 
not only to compile yet another dataset with completely new data, but rather 
to: \emph{(i)} create a common framework to test tracking methods on;  
%
\emph{(ii)} gather existing and new challenging sequences 
with very different characteristics (frame rate, 
pedestrian density, illumination, or point of view) in order to challenge 
researchers to develop more general tracking methods that can deal with all 
types of sequences.
An overview of the characteristics of the 2015 and 2016 releases is shown in Table \ref{tab:datasets}. More detailed information on each individual sequence can be found in the supplementary material.

\myparagraph{\MOTOLD sequences.}
Our first release consisted of 22 sequences, 11 each for 
training and testing. The test data contains over 10 minutes 
of footage and 61440 annotated bounding boxes, making it hard to overtune on such a large amount of data, one of the 
benchmark's major strengths. 
Among the 22 sequences, there are six new challenging high-resolution videos, four filmed from a static 
and two from a moving camera held at pedestrian's height. Three of them are particularly difficult: a night sequence from a moving camera and two outdoor sequences with a high density of pedestrians. The moving 
camera together with the low illumination creates a lot of motion blur, 
making this sequence extremely challenging. %


\myparagraph{\MOTNEW sequences.}
For the second data release, we focused on two aspects: \emph{(i)} increasing the difficulty of the challenge, \eg by having scenarios with a 3 times higher mean density of pedestrians; and \emph{(ii)} improving annotations by following a strict annotation protocol that also included several classes aside from the class {\it pedestrian}. 
%
As can be seen in Table \ref{tab:datasets}, the new data contains almost three times more bounding boxes for training and testing compared to \MOTOLD. Most sequences are filmed in high resolution. 
Aside from pedestrians, the annotations also include other classes like vehicles, bicycles, \etc in order to provide contextual information for methods to exploit.

\subsection{Annotation Protocol and Ground Truth}

For \MOTOLD, most of the ground truth bounding boxes were public. New annotations were provided for the six new sequences.
One weakness of this first release was that the annotation protocol was not consistent across all sequences.
In order to improve the above shortcoming, the annotations for \emph{all} \MOTNEW sequences have been carried out by qualified researchers from scratch following a strict protocol, and finally double-checked to ensure highest annotation accuracy. Not only pedestrians were annotated, but also vehicles, sitting people, occluding objects, as well as other significant object classes. With this fine-grained level of annotation it is possible to accurately compute the degree of occlusion and cropping of all bounding boxes, which is also provided with the benchmark. The detailed annotation protocol can be found in the supplementary material.

Before the release of our benchmark, and with the exception of KITTI \cite{Geiger:2012:CVPR}, it was common in pedestrian tracking to evaluate on a few handpicked sequences for which ground truth was known. In our setup, none of the test ground truth is public, preventing methods from overfitting to a particular scenario, with the hope to make trackers more general.

\subsection{Detections}
\label{sec:detections}


\definecolor{detcolor}{rgb}{0.4472,    0.4472,    0.8944}
\begin{figure}
\centering
   \includegraphics[width=.48\linewidth]{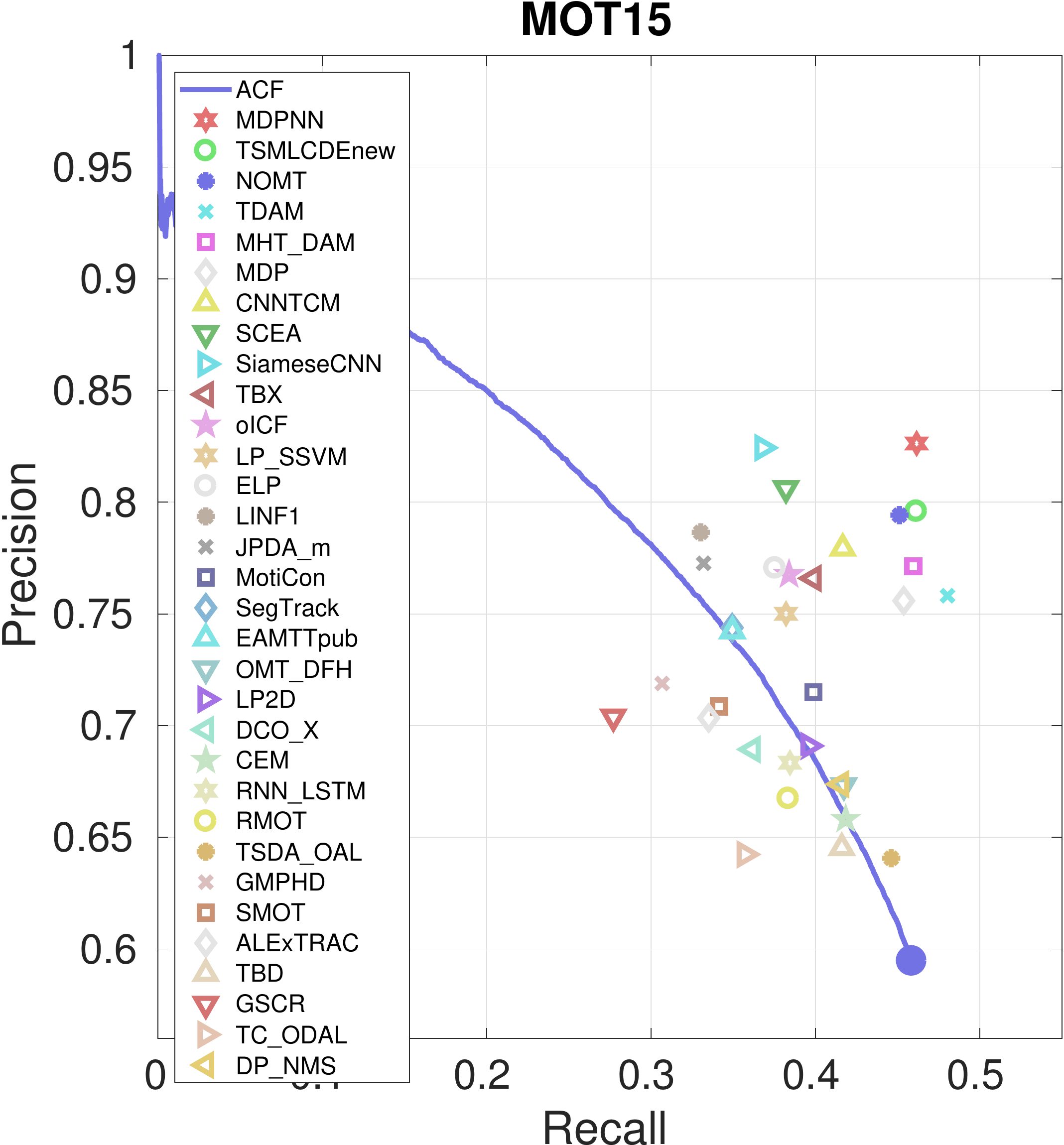}\hfill
     \includegraphics[width=.48\linewidth]{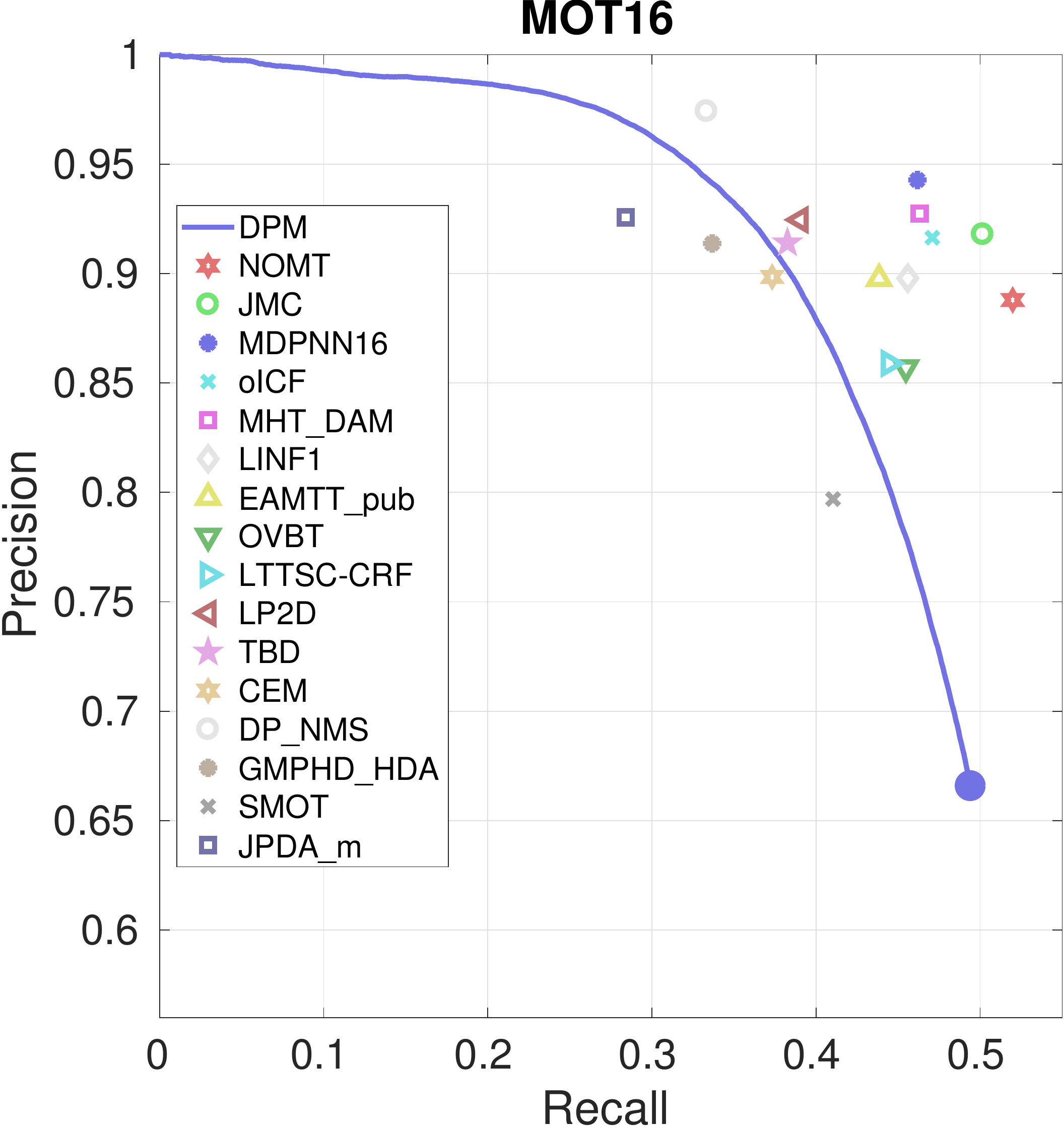}
  \caption{Precision and recall for \MOTOLD (left) and \MOTNEW. The accompanying detector (ACF\cite{Dollar:2014:PAMI} for \MOTOLD, DPM \cite{Felzenszwalb:2010:PAMI} for \MOTNEW), applied on each frame of the sequence, is plotted as a solid line. 
  We consider all annotated pedestrians, even fully occluded ones, explaining the relatively low recall of both detectors.
  Trackers typically do not provide a confidence and are thus plotted with dots. Note that, when compared to the full detector set (\textcolor{detcolor}{$\bullet$}), most trackers are only able to improve precision, but struggle to reduce the number of missed targets.
  }
 \label{fig:detectors}
 \vspace{-0.5em}
\end{figure}

Arguably, the detector plays a crucial role for all tracking-by-detection methods.
To focus the benchmark on the tracking task, we provide detections for all images. For \MOTOLD, we used the Aggregated Channel Feature (ACF) detector 
\cite{Dollar:2014:PAMI}, with default parameters and trained on the INRIA dataset \cite{Dalal:2005:CVPR}.
For \MOTNEW, we tested several state-of-the-art detectors, including Fast R-CNN \cite{girschickcvpr2015}. However, when using the pre-trained, off-the-shelf model, the deformable part-based model (DPM) v5 \cite{Felzenszwalb:2010:PAMI} outperformed the other detectors in the task of pedestrian detection on our dataset. This is consistent with the observations made in \cite{girschickcvpr2015}, stating that the out-of-the-box R-CNN outperforms DPM in detecting all object classes except for the class \emph{person}.
All evaluated tracking methods use the same set of detections as their input: ACF detections for \MOTOLD and DPM detections for \MOTNEW;
see \Fig\ref{fig:detectors} for precision-recall curves.


\subsection{Submission and Evaluation}
To limit the effect of overfitting, we follow certain submission guidelines similar to other existing benchmarks~\cite{Geiger:2012:CVPR,Andriluka:2014:CVPR,Russakovsky:2014:ImageNET}. We limit the total number of submissions to four for one particular approach, and we enforce a minimum 72-hour gap between submissions. This has proven to be a good and effective practice to prevent participants from overly tuning their parameters to the test set.

A critical point with any dataset is how to measure the performance of 
the algorithms. We will discuss and analyze different existing metrics in \Sec~\ref{sec:futuremetrics}. A clear advantage of the proposed benchmark is that all metrics are computed in a centralized way and with the same exact ground truth, allowing a fair comparison of different tracking approaches.

\begin{figure*}[t]
\centering
\vspace{1.5em}
\begin{overpic}[width=0.24\linewidth]{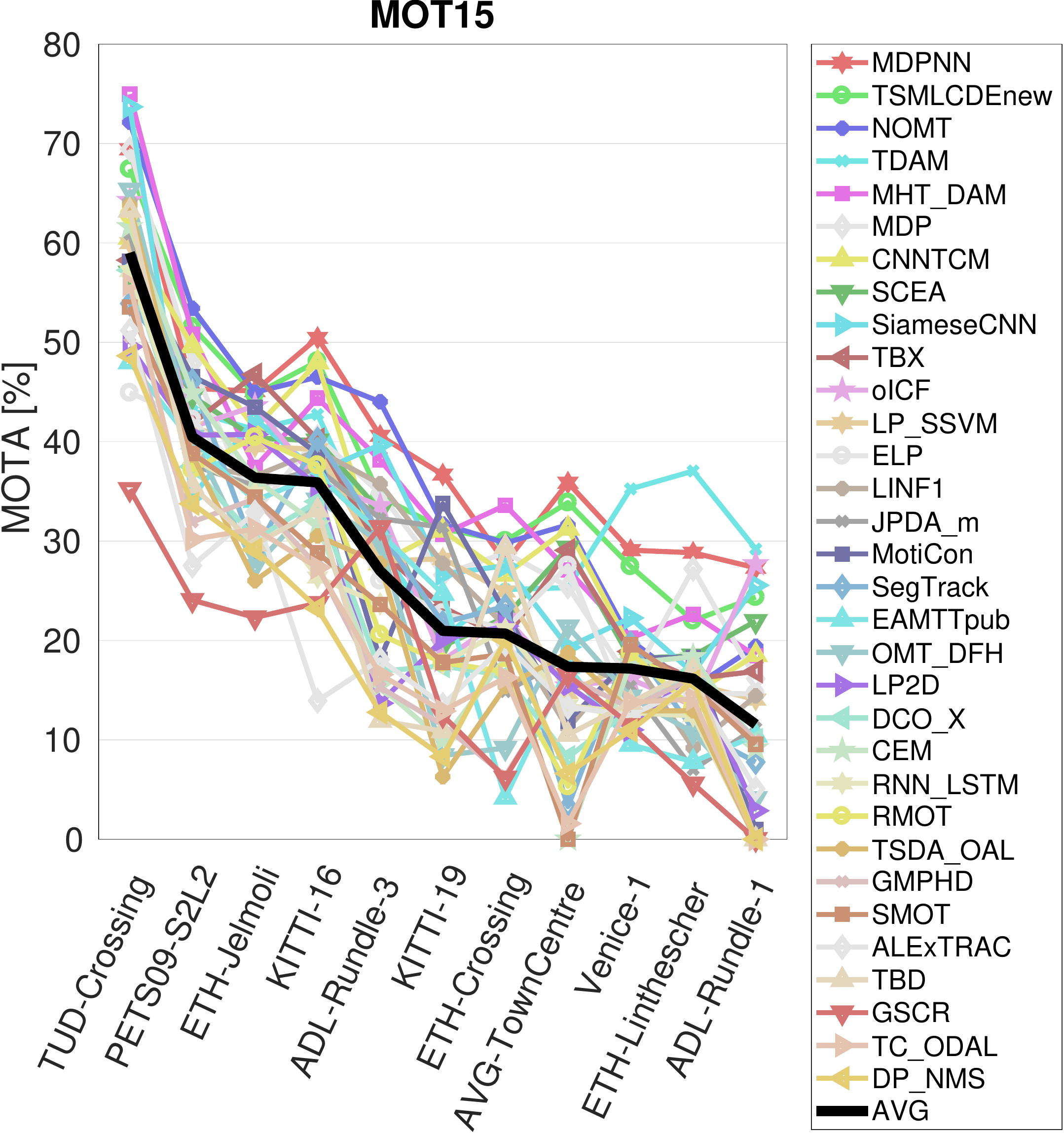}
 \put (45,105) {\footnotesize \textsf{\textbf{Per-sequence performance}}}
\end{overpic}
  \includegraphics[width=.24\linewidth]{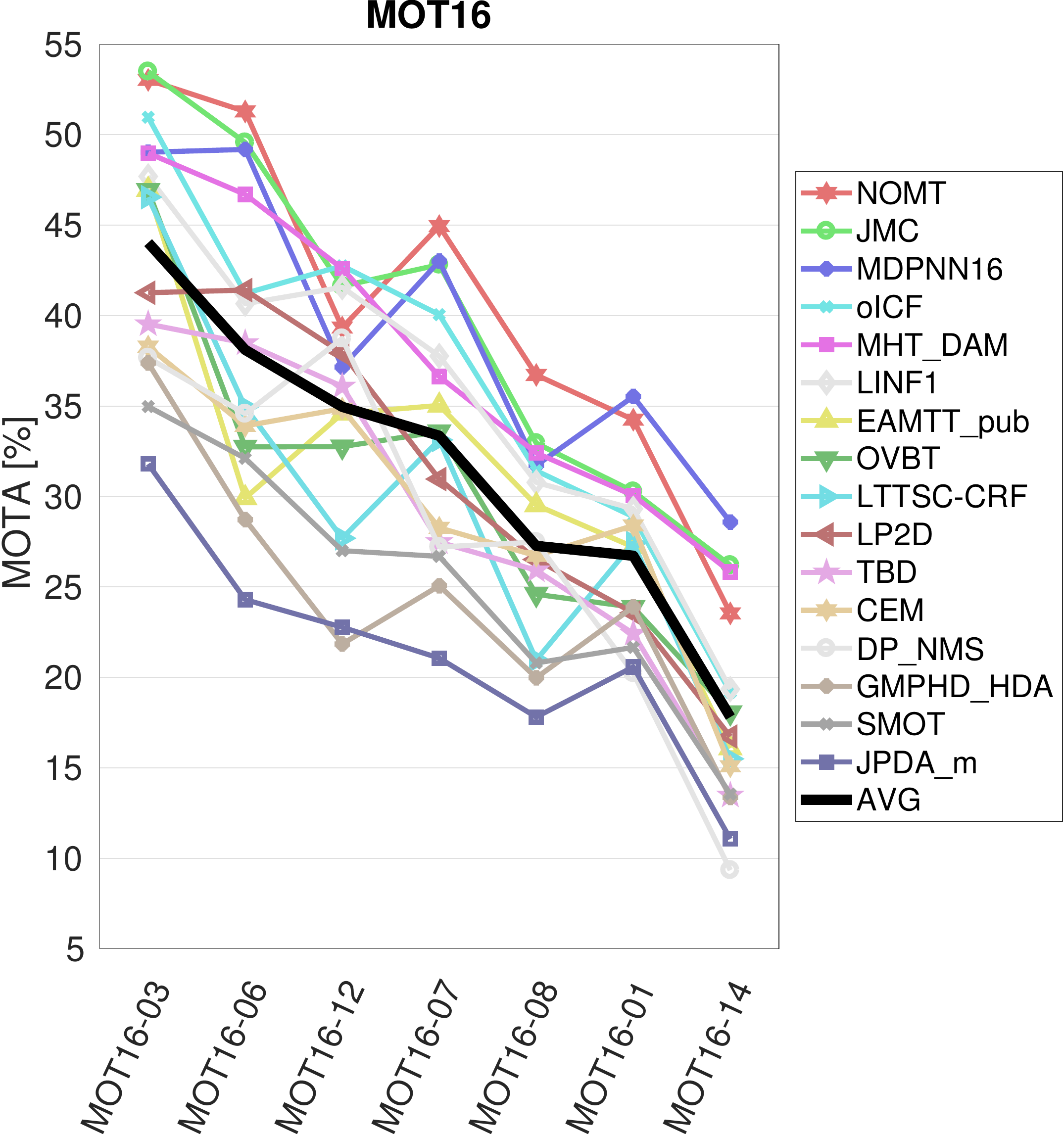}
  \hfill
\begin{overpic}
[width=0.23\linewidth]{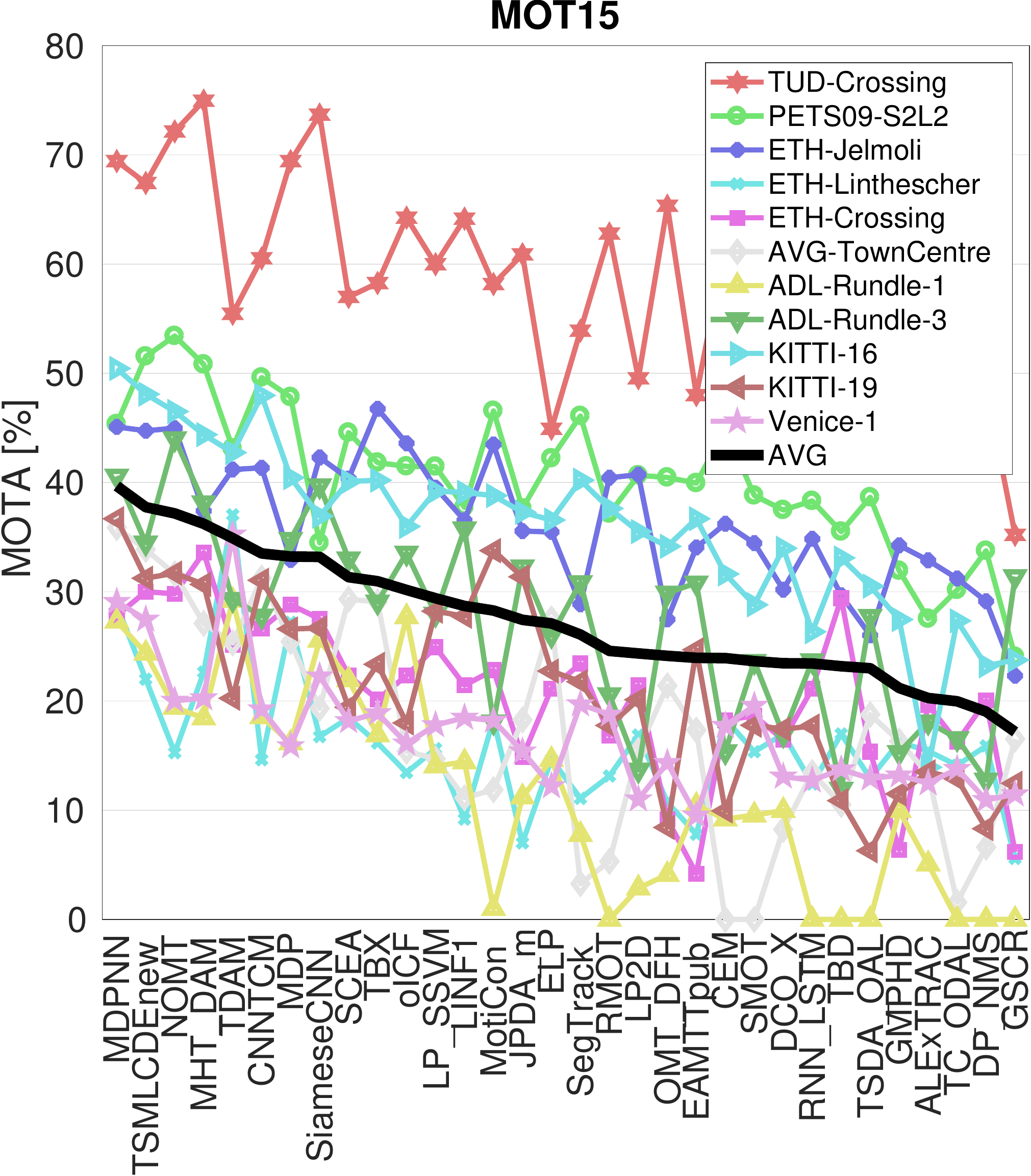}
 \put (55,105) {\footnotesize \textsf{\textbf{Per-tracker performance}}}
\end{overpic}  
  \includegraphics[width=.23\linewidth]{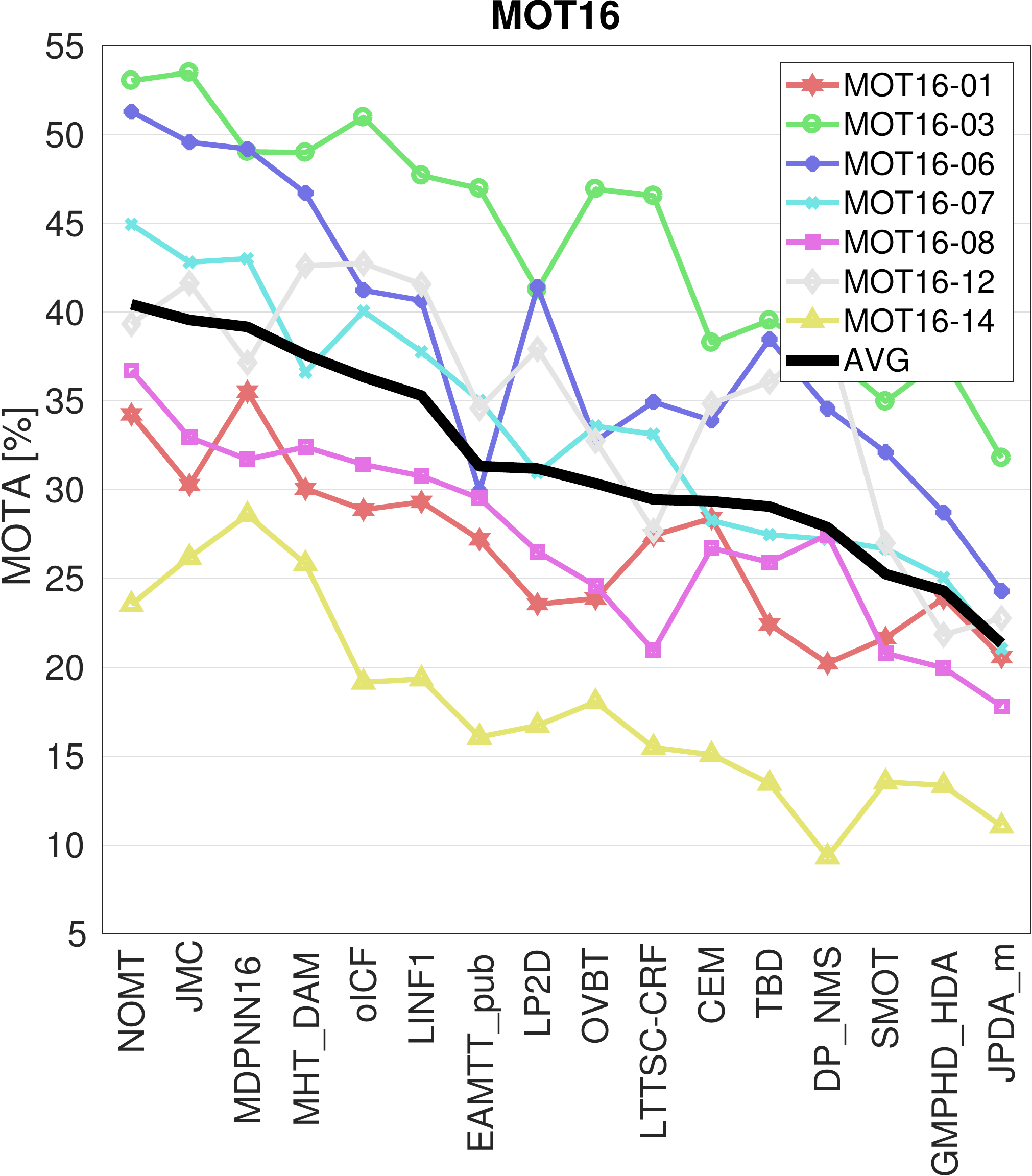}
  \caption{Graphical overview of all submissions. The entries are ordered from easiest sequence / best performing method, to hardest sequence / poorest performance, respectively. The mean performance across all sequences / submissions is depicted with a thick black line.}
 \label{fig:results-detailed}
 \vspace{-0.5em}
\end{figure*}

\section{Analysis of State-of-the-Art Trackers}
\label{sec:experiments}
We now present and analyze the results of all submissions, and highlight certain trends of the community. We consider all valid submissions to the benchmark that were published before March 1st, 2017 and used the provided set of detections. The total number of tracking results was 48, 32 of which were tested on \MOTOLD, and the remaining 16 on \MOTNEW. Note that 13 methods were tested on both datasets. Also note that a small subset of submissions\footnote{The methods DP$\_$NMS, TC$\_$ODAL, TBD, SMOT, CEM, DCO$\_$X, LP2D were taken as baselines for the benchmark.} was done by the benchmark organizers and not by the original authors of the respective method.
Results for \MOTNEW are summarized in Table~\ref{tab:mot16}. 
\Fig~\ref{fig:results-detailed} provides a graphical overview of performance as measured by MOTA for all submissions on both datasets; see the supplementary material for more details.




\begin{figure*}
\centering
   \includegraphics[width=0.24\linewidth]{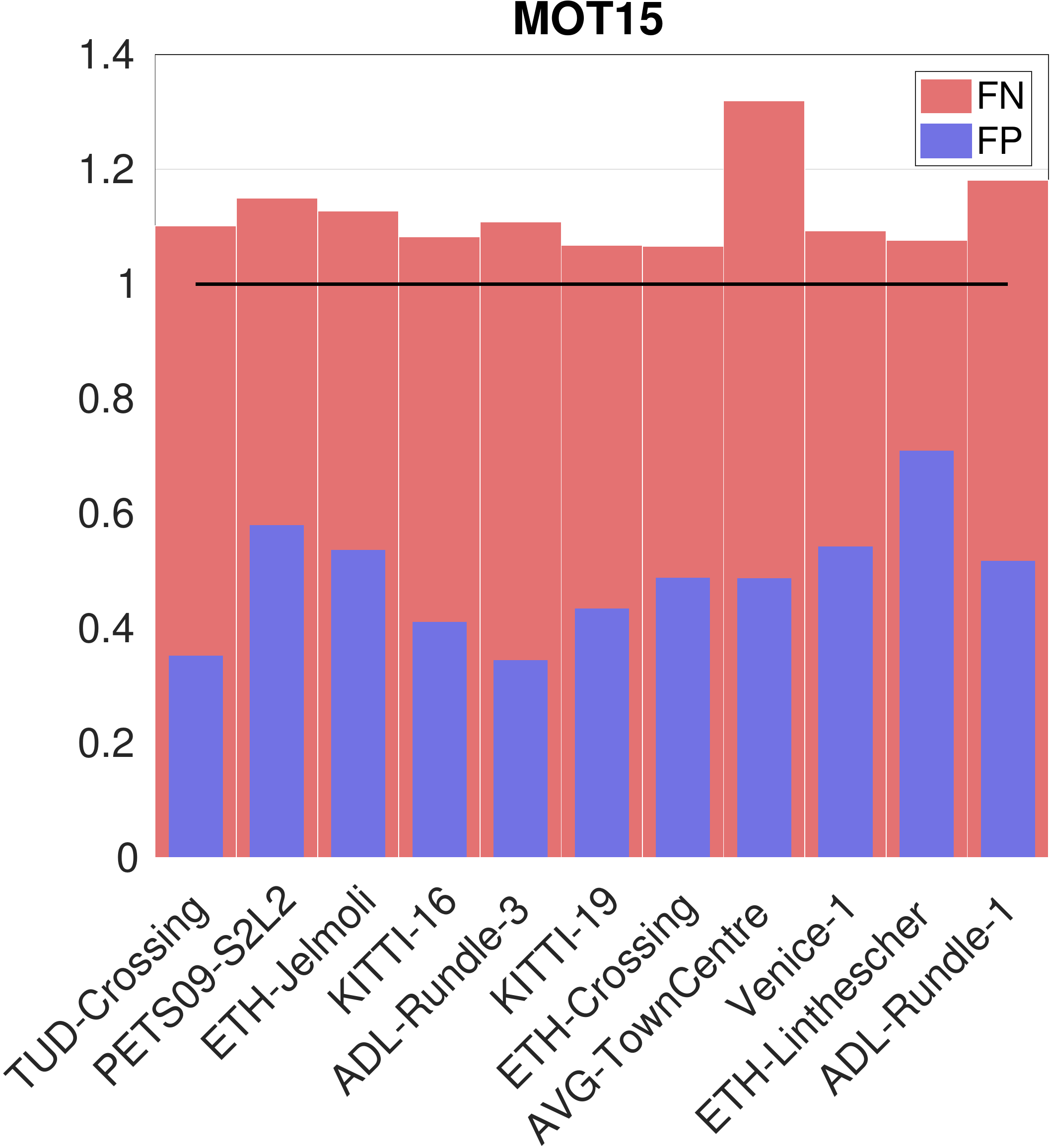}     
  \includegraphics[width=0.225\linewidth]{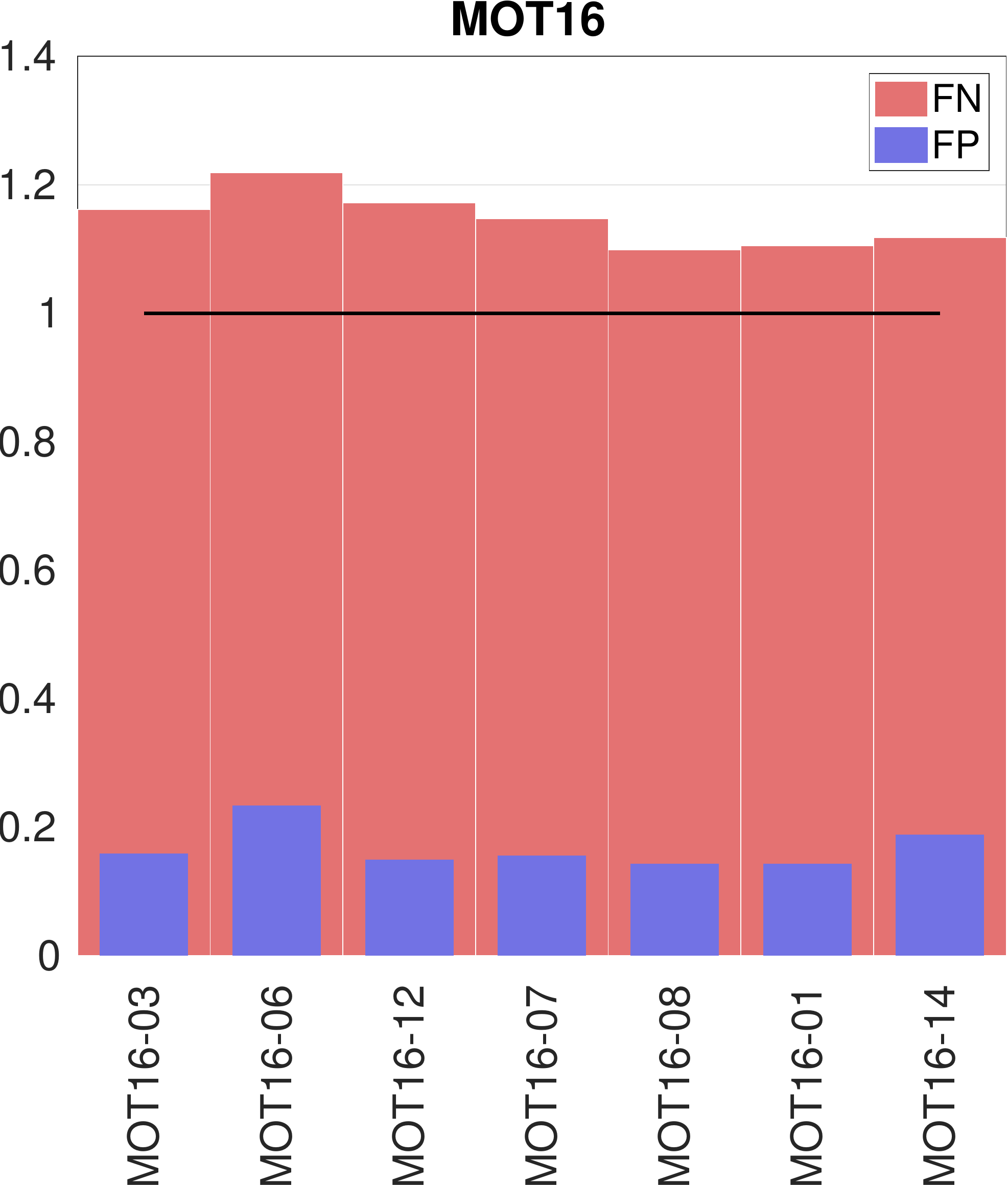}
  \hfill
  \includegraphics[width=0.225\linewidth]{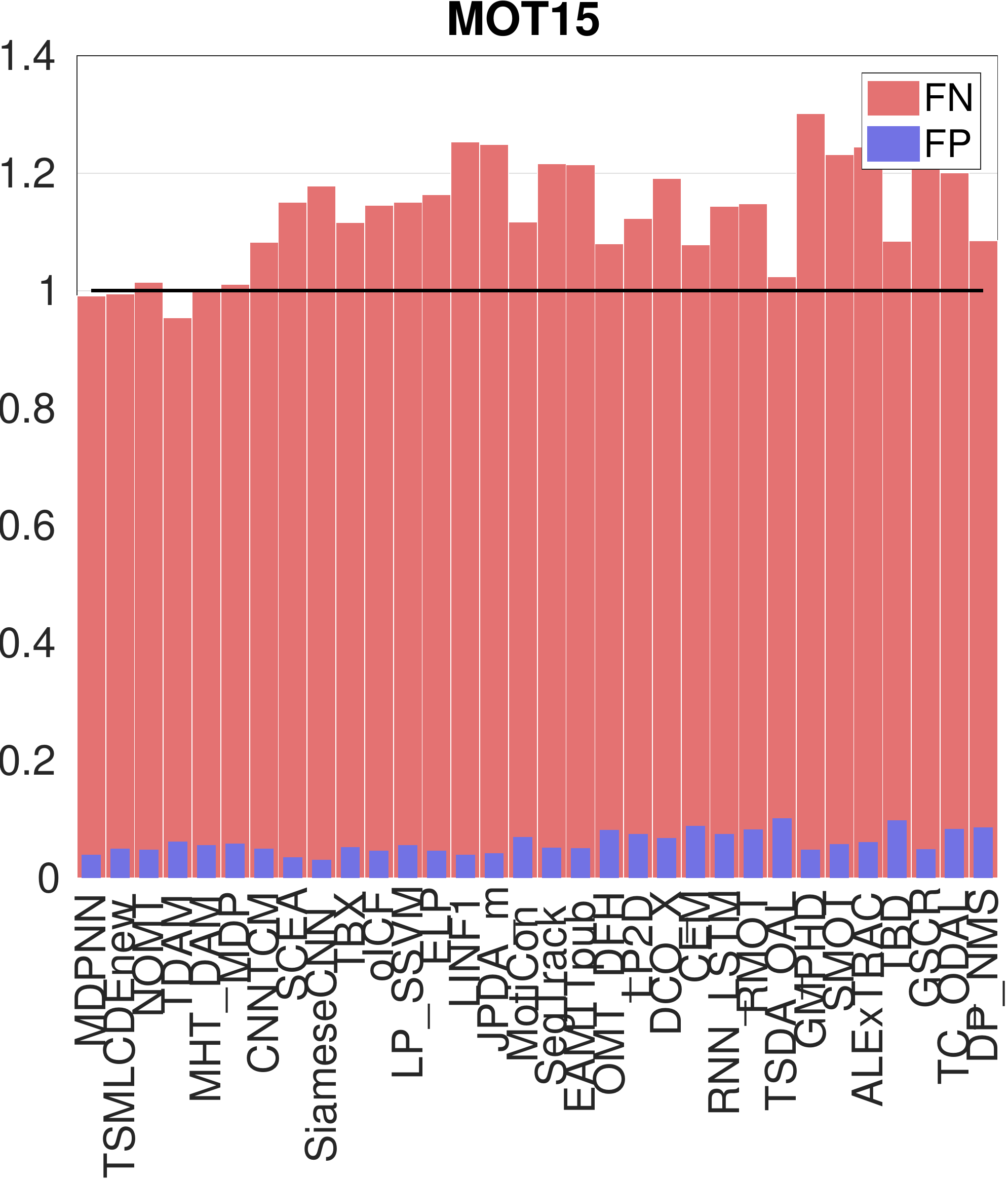}
     \includegraphics[width=0.225\linewidth]{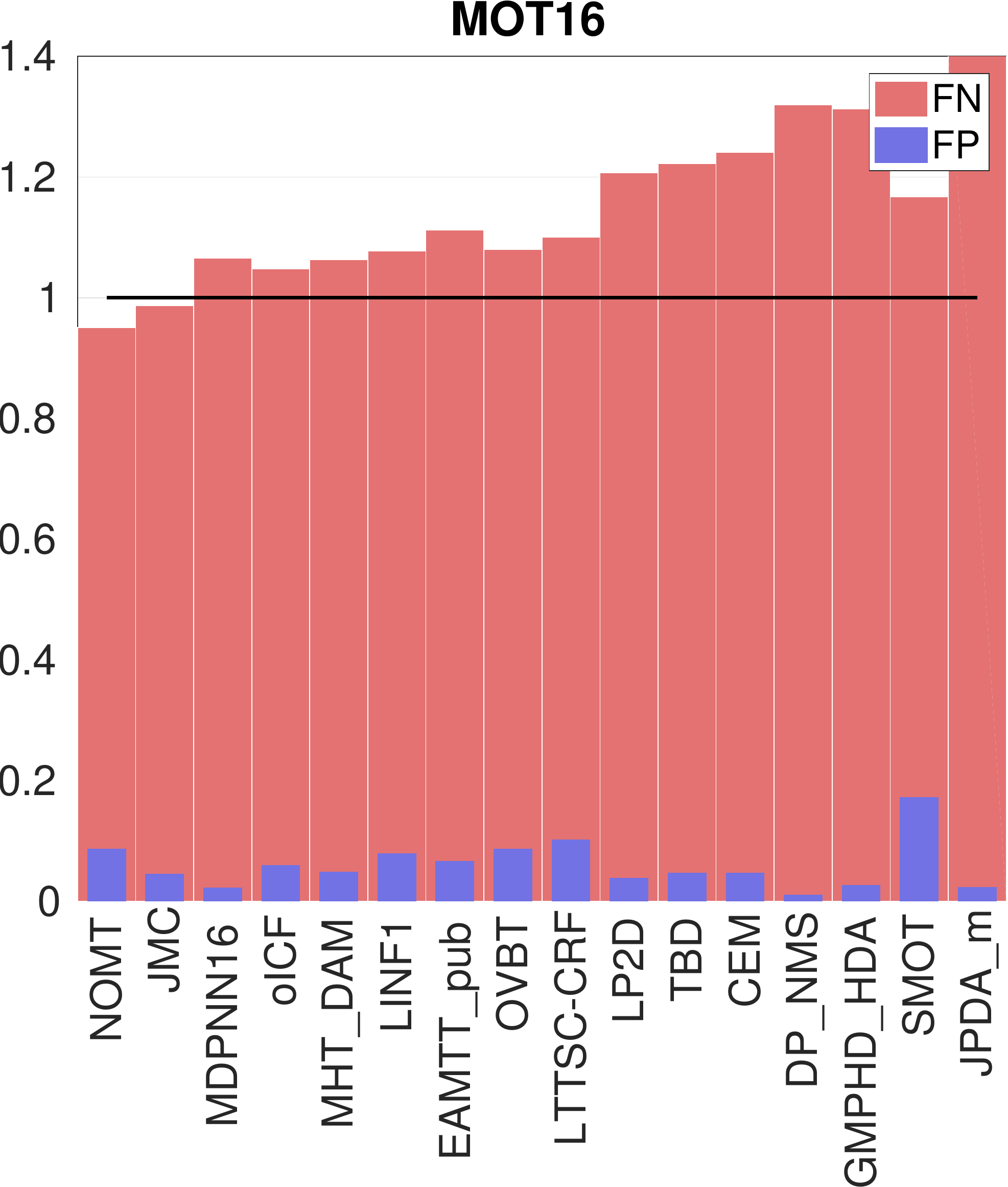}
  \caption{Detailed error analysis. The plots show the error ratios for trackers \wrt detector (taken at the lowest confidence threshold), for two types of errors: false positives (FP) and false negatives (FN). Values above $1$ indicate a higher error count for trackers than for detectors. Note that most trackers concentrate on removing false alarms provided by the detector at the cost of eliminating a few true positives, indicated by the higher FN count.}
 \label{fig:FNFP}
\end{figure*}

\subsection{Trends in Tracking}
We first turn our attention to the main trends in the multi-object tracking literature. Looking at Table \ref{tab:mot16}, we can quickly distinguish a set of 6 top-performing trackers~\cite{Choi:2015:ICCV,tangBMTT2016,SadeghianArxiv2017,kieritzAVSS2016,Kim:2015:ICCV,FagotECCV2016}, with MOTA above 40\% and more than 10\% Mostly Tracked trajectories.
What distinguishes those trackers from the rest?

\newcolumntype{L}[1]{>{\raggedright\let\newline\\\arraybackslash\hspace{0pt}}m{#1}}
\newcolumntype{C}[1]{>{\centering\let\newline\\\arraybackslash\hspace{0pt}}m{#1}}
\newcolumntype{R}[1]{>{\raggedleft\let\newline\\\arraybackslash\hspace{0pt}}m{#1}}


\begin{table*}[h!]
  \caption{The \MOTNEW leaderboard. Performance of several trackers according to different metrics.}
  \label{tab:mot16}
  \smallskip
\centering
\begin{tabular*}{\linewidth}{@{\extracolsep{\stretch{1}}}l rrrrrrrrr @{}}
\toprule
  Method & MOTA & MOTP & FAF & MT & ML & FP & FN & IDsw & Frag \\
  \midrule
 NOMT \cite{Choi:2015:ICCV} & 46.4$\pm$9.9	&76.6&	1.6&	18.3&	41.4&	9753&	87565&	359 (6.9)&	504 (9.7)\\

 JMC \cite{tangBMTT2016} & 46.3$\pm$9.0	& 75.7& 	1.1	& 15.5& 	39.7& 	6373	& 90914& 	657 (13.1)	& 1114 (22.2)\\
 
  MDPNN16 \cite{SadeghianArxiv2017} & 43.8$\pm$7.3	& 75.5	& 0.6	& 12.4& 	40.7& 	3501& 	98193& 	723 (15.7)	& 2036 (44.1)\\

  oICF \cite{kieritzAVSS2016} & 43.2$\pm$10.2	& 74.3	& 1.1	& 11.3& 48.5& 	6651& 	96515	& 381 (8.1)& 	1404 (29.8)
 \\
 
 MHT\_DAM \cite{Kim:2015:ICCV} & 42.9$\pm$8.9&	76.6&	1.0&	13.6&	46.9&	5668&	97919	&499 (10.8)&	659 (14.2)
 \\

 LINF1 \cite{FagotECCV2016} & 41.0$\pm$9.5&	74.8	&1.3&	11.6&	51.3&	7896	&99224&	430 (9.4)	&963 (21.1)
\\
 
      EAMTT\_pub \cite{sanchezbmtt2016} & 38.8$\pm$8.5&75.1&	1.4	&7.9&	49.1&	8114&	102452	&965 (22.0)	&1657 (37.8)
\\
 
 OVBT \cite{BanBMTT2016} & 38.4$\pm$8.8&	75.4 &	1.9	 &7.5 &	47.3 &	11517 &	99463 &	1321 (29.1)	 &2140 (47.1)
\\ 
 LTTSC-CRF \cite{LeBMTT2016} & 37.6$\pm$9.9	&75.9	&2.0	&9.6&	55.2&	11969&	101343	&481 (10.8)&	1012 (22.8)
\\ 
  LP2D \cite{lealiccv2011} & 35.7$\pm$10.1&	75.8&	0.9&	8.7&	50.7&	5084	&111163&	915 (23.4)&	1264 (32.4)\\
 
 TBD \cite {Geiger:2014:PAMI} & 33.7$\pm$9.2	&76.5&	1.0&	7.2&	54.2&	5804&	112587	&2418 (63.2)	&2252 (58.9)
\\
 
            CEM \cite{Milan:2014:PAMI} & 33.2$\pm$7.9	& 75.8	& 1.2& 	7.8& 	54.4& 	6837& 	114322& 	642 (17.2)	& 731 (19.6)
\\
 
               DP\_NMS \cite{Pirsiavash:2011:CVPR} & 32.2$\pm$9.8&	76.4	&0.2	&5.4&	62.1&	1123&	121579&	972 (29.2)&	944 (28.3)
\\
 
   GMPHD\_HDA \cite{SongICCE2016} & 30.5$\pm$6.9&	75.4&	0.9	&4.6&	59.7&	5169	&120970	&539 (16.0)&	731 (21.7)
\\
 
   SMOT \cite{Dicle:2013:ICCV} & 29.7$\pm$7.3&	75.2	&2.9	&5.3&	47.7&	17426&	107552&	3108 (75.8)&	4483 (109.3)
 \\
 
   JPDA\_m \cite{Rezatofighi:2015:ICCV} & 26.2$\pm$6.1&	76.3	&0.6	&4.1&	67.5&	3689	&130549	&365 (12.9)&	638 (22.5)
\\
 
\bottomrule
\end{tabular*}
\end{table*}

\myparagraph{Data association.}
Before 2015, the community mainly focused on finding strong, preferably globally optimal, methods to solve the data association problem. 
The task of linking detections into a consistent set of trajectories was often cast as a graphical model and solved with k-shortest paths in DP$\_$NMS \cite{Pirsiavash:2011:CVPR}, as a Linear Program solved with the simplex algorithm in LP2D \cite{lealiccv2011}, in a Conditional Random Field as in DCO$\_X$ \cite{milanTPAMI2016}, SegTrack~\cite{Milan:2015:CVPR} and LTTSC-CRF \cite{LeBMTT2016}, or as a variational Bayesian model as in OVBT \cite{BanBMTT2016}, to name a few.
A lot of attention was also given to motion models such as SMOT \cite{Dicle:2013:ICCV}, CEM \cite{Milan:2014:PAMI}, or MotiCon \cite{lealcvpr2014}.
The pairwise costs for matching two detections were based on either simple distances or weak appearance models.
These methods achieve around 38\% MOTA on \MOTNEW and 25\% on \MOTOLD, which is 10\% below current state of the art.

\myparagraph{Affinity and appearance.}
More recently, the attention shifted towards building robust pairwise similarity costs, mostly based on strong appearance cues.
This shift is clearly reflected in an increase in tracker performance, and the ability for trackers to handle more complex scenarios. 
The top performing methods use sparse appearance models in LINF1 \cite{FagotECCV2016}, online appearance updates in MHT$\_$DAM \cite{Kim:2015:ICCV}, integral channel feature appearance models in oICF\cite{kieritzAVSS2016}, and aggregated local flow of long-term interest point trajectories in NOMT \cite{Choi:2015:ICCV} to improve detection affinity.
Deep learning has also had an impact on tracking, however its application to the problem at hand remains rather sparse. One example is MDPNN16~\cite{SadeghianArxiv2017}, which leverages Recurrent Neural Networks in order to encode appearance, motion, and interactions. JMC\cite{tangBMTT2016} uses deep matching to improve the affinity measure. 

In summary, the main common component of top performing methods are strong affinity models. We believe this to be one of the key aspects to be addressed to further improve performance; we expect to see many more approaches that attempt to accomplish this using deep learning.


\newcommand{\void}{---}
\newcommand{\fillme}{\textcolor{red}{\bf ??}}
\begin{table*}[tb]
\centering
\footnotesize
\caption{\MOTNEW trackers and their characteristics.}
\begin{tabular}{@{} l  c c c c c c c c@{}} 
\toprule
  Method& Box-box Affinity  & Appearance & Optimization & Extra Inputs & Online \\ \midrule
 NOMT \cite{Choi:2015:ICCV}   &  Interest Point Trajectories &  \cmark &  CRF & Optical flow & \xmark\\
 JMC \cite{tangBMTT2016}      & DeepMatching & \cmark & Multicut & Non-NMS dets & \xmark  \\
 MDPNN16 \cite{SadeghianArxiv2017} & RNN (motion, appearance, interactions) &  \cmark & Markov Decision Process & \void & \cmark\\
 oICF \cite{kieritzAVSS2016} & Motion model + MIL on appearance & \cmark  & Kalman filter & \void & \cmark \\
 MHT\_DAM \cite{Kim:2015:ICCV} & Regression classifier appearance & \cmark & Multiple Hypothesis & \void & \xmark \\
 LINF1 \cite{FagotECCV2016} & Sparse representations appearance & \cmark & MCMC  & \void & \xmark \\
 EAMTTpub \cite{sanchezbmtt2016} & 2D distances & \xmark & Particle Filter & Non-NMS dets &  \cmark \\
 OVBT \cite{BanBMTT2016} & Dynamics from flow & \cmark & Variational EM & Optical flow & \cmark\\
 LTTSC-CRF \cite{LeBMTT2016} & SURF & \cmark & CRF & SURF &\xmark \\
 LP2D \cite{lealiccv2011} & 2D image distances, IoU & \xmark & Global, LP & \void & \xmark\\
 TBD \cite {Geiger:2014:PAMI} & IoU + NCC & \cmark & Hungarian algorithm & \void & \xmark\\
 CEM \cite{Milan:2014:PAMI} & 2D velocity difference & \xmark & L-BFGS + greedy sampling & \void &\xmark\\
 DP\_NMS \cite{Pirsiavash:2011:CVPR} & 2D image distances & \xmark & k-shortest paths & \void & \xmark\\
 GMPHD\_HDA \cite{SongICCE2016} & HoG similarity, color histogram & \cmark & Gaussian mixture PHD filter & HoG & \cmark \\
 SMOT \cite{Dicle:2013:ICCV} & Target dynamics & \xmark &   Hankel Total Least Squares & \void & \xmark \\
 JPDA\_m \cite{Rezatofighi:2015:ICCV} & Mahalanobis distance & \xmark & LP & --- & \xmark \\
 \bottomrule
\end{tabular}
\label{tab:characteristics}
\vspace{-0.5em}
\end{table*}


\subsection{Error Analysis}
\label{sec:error-analysis}


We now take a closer look at the most common errors made by the tracking approaches. 
In \Fig~\ref{fig:FNFP}, we show the number of false negatives (FN, blue) and false positives (FP, red) created by the trackers on average with respect to the number of FN/FP of the input detector. A ratio below $1$ indicates that the trackers have improved in terms of FN/FP over the detector, while values above $1$ mean a performance decrease. 
On the left, we see the performance for each sequence averaged over trackers; the sequences are ordered by decreasing MOTA.
On the right, we show the performance for each tracker averaged over sequences.

We observe that while trackers are good at reducing FPs, most barely reduce the number of false negatives. Many of them even increase it by 20-30\%. 
This is contrary to the common wisdom that trackers are good at filling the gaps between detections and creating full trajectories. 
Moreover, we can see a direct correlation between the FN and tracker performance. This is because there is a much larger number of FN than FP (detailed values can be found in the supplementary material), hence their weight on the MOTA value is much larger. 
One important question that arises is that if FNs are so important, why do trackers not focus more attention on reducing them?



One possible way of tackling the problem would be to get only very confident detection and create a tracker that focuses on filling the gaps to obtain long trajectories.
In order to find if this could be a valid strategy, we computed the percentage of trajectories covered by the detections (determined by 50\% IoU score). Taking all detections into account, as much as 18\% of the trajectories are not covered by any detection. What is even more surprising is that if we drop the 10\% detections with the lowest confidence, the number of completely uncovered tracks goes up to 55\%.
Of course, these trajectories will never be recovered by any tracker because there is no remaining evidence at all for that pedestrian.
Thereby, this strategy would lead to more FNs in the end than current strategies that focus on better recovering the feasible trajectories.




\section{Predicting Performance}
\label{sec:futuresequences}

In this section, we investigate the question whether it is possible to predict how currently available tracking methods would perform on a particular video. This would allow us to create a `super tracker' by choosing the best approach each time we are confronted with a new video. We carry out this analysis at two granularity levels: \emph{(a)} at a sequence level, and \emph{(b)} at the level of short video snippets. For the latter, the entire dataset is divided into temporally overlapping 50-frame long fragments, with a stride of 25 frames. 
The benefit of this fine-grained analysis is twofold. On one hand, it allows us to precisely pinpoint the strengths and weaknesses of current methods for a particular situation without averaging over the entire video sequence. On the other hand, this approach generates enough data for training a classifier.

\myparagraph{Fine-grained performance analysis.}
Results of this fine-grained evaluation are illustrated in \Figs~\ref{fig:snippets} and \ref{fig:complexity-snippets}.
\Fig~\ref{fig:snippets} depicts an example of this evaluation on the KITTI-19 sequence. We only show a subset of all trackers to maintain readability. Please refer to the supplemental material for more examples. \Fig~\ref{fig:complexity-snippets} shows the same evaluation for \emph{all} trackers and \emph{all} sequences as a heatmap. On the right, three example frames for the locally highest, locally lowest, and locally intermediate MOTA averaged across all submissions are shown. It is interesting to note that for both \MOTOLD and \MOTNEW, two out of the three examples are picked from the same sequence, which demonstrates the performance difference within one video sequence. Looking at the two hardest examples, we can observe two similarities. First, they both fall within the `driving' scene scenario, which is typically more difficult due to camera motion. Second, these fragments contain relatively few true targets such that the MOTA is largely dominated by the false positive tracks which are not suppressed. Please see the supplemental video for this and more examples.

\begin{figure}[t]
\centering
  \includegraphics[width=1\linewidth]{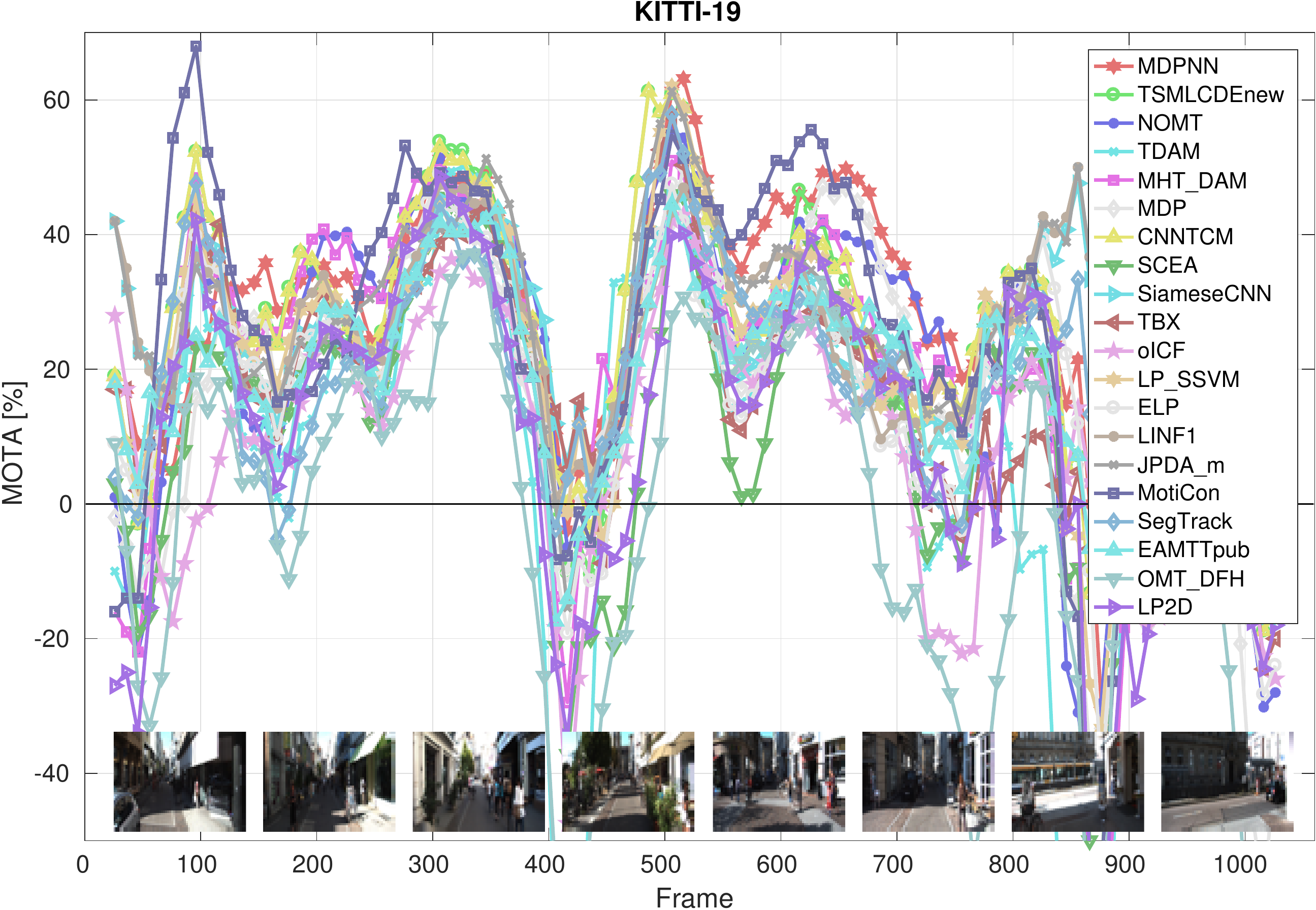}
\caption{An example of fine-grained analysis for the KITTI-19 sequence. Each line represents the \emph{local} performance of a tracker, measured by MOTA within a 50-frame (roughly 3 seconds long) segment. Note the extreme within-sequence variation.}
 \label{fig:snippets}
 \vspace{-0.5em}
\end{figure}


\begin{figure*}[t]
\centering
  \includegraphics[width=1\linewidth]{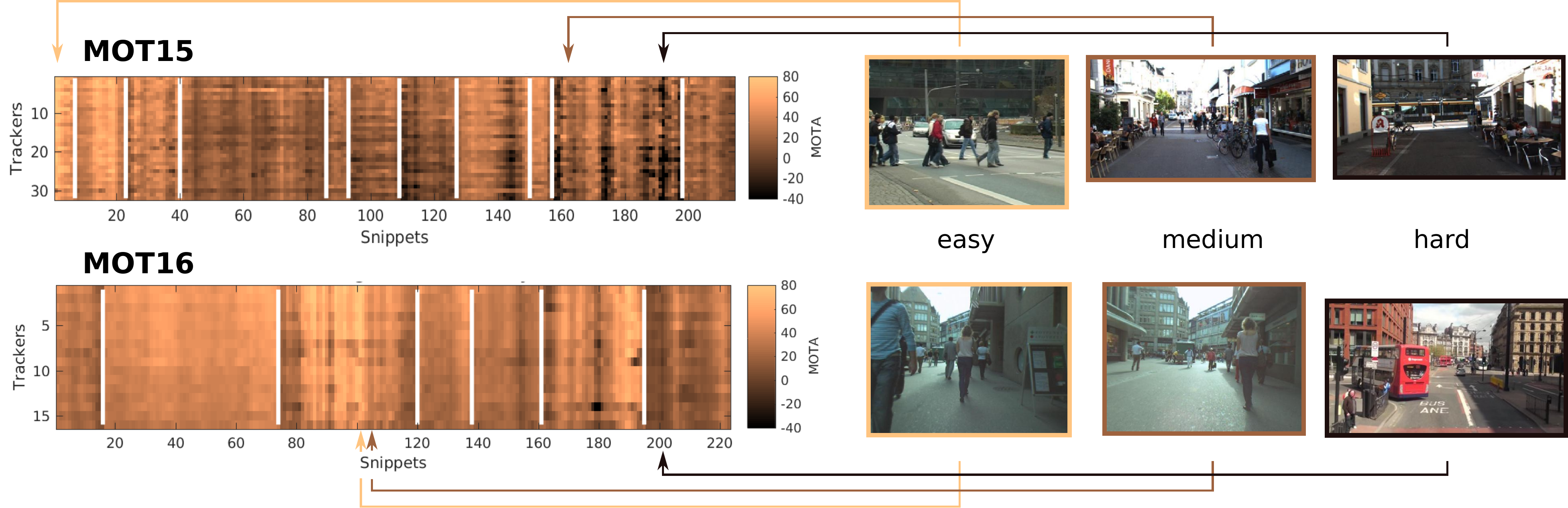}  
  \caption{A fine-grained video complexity analysis for \MOTOLD (top) and \MOTNEW (bottom) for all trackers. Each row shows a performance heatmap for all trackers, measured by MOTA in 50-frame snippets. The three frames on the right represent the easiest, average, and most difficult video section, averaged over all trackers. Note the high within sequence performance difference in both cases.}
 \label{fig:complexity-snippets}
 \vspace{-0.5em}
\end{figure*}

\begin{table}[b]
\vspace{-0.5em}
\caption{Combining tracker output. The results reflect MOTA of the state-of-the-art method (best) and a modified version by considering the top-3 performing submissions. See text for details.}
\vspace{0.1cm}
\label{tab:super-tracker}
\centering
\small
\begin{tabular}{@{}l  ccc  cc@{}}
\toprule
 &\multicolumn{3}{c }{per-sequence} &  \multicolumn{2}{c}{fine-grained}\\
 \cmidrule(l{2pt}r{2pt}){2-4}  \cmidrule(l{2pt}r{2pt}){5-6} 
Dataset &  best & predict & oracle  & predict & oracle\\
\midrule
\MOTOLD & 36.6 & 36.7 & 40.5 & 39.5 & 42.1 \\
\MOTNEW &46.4 & 46.2 & 47.4 &  46.0 & 48.3 \\

\bottomrule
\end{tabular}
\end{table}
For both levels of granularity we performed the following experiment. We took each sequence (or a shorter video fragment), computed several features and trained a linear multi-class SVM using cross validation to predict which of the top-three trackers produces the best result on that sequence. We used 7 different features:
min/max scale ratio of the top 50\% detections,
the mean number of detections per frame,
the mean detection confidence,
the mean mean and mean max flow magnitudes of entire frames, and those of only the non-person regions.
The first three features represent scene geometry and saliency, and indicate whether people are easy to be detected. The optic flow features estimate the presence of camera motion.

The prediction for each temporal segment was made using the split that did not contain this data point in its training set. To obtain the final tracking result, the individual outputs are simply concatenated without sophisticated association schemes. We also provide an upper bound on the best possible performance gain by selecting the optimal tracker result for each video segment.

The results are summarized in \Tab~\ref{tab:super-tracker}. Interestingly, even with the oracle prediction for each segment, the overall improvement on MOTA compared to the currently best performing method is moderate. It is $5.5$ percentage points for \MOTOLD and merely $2.9$ percentage points for \MOTNEW. 
Using the trained SVMs prediction, we can only slightly improve on \MOTOLD, however the classifier is too error-prone on the \MOTNEW dataset, even leading to a minor decay. Considering the complexity analysis in \Fig~\ref{fig:complexity-snippets}, this is not entirely surprising. \MOTOLD contains a much higher diversity of results for a particular video fragment, while \MOTNEW is fairly homogeneous in comparison. This makes it hard to predict the best method, but also to improve the overall performance using the oracle prediction.




\section{Analysis of the Evaluation Metrics}
\label{sec:futuremetrics}

One of the key aspects of any benchmark is the evaluation protocol.
In the case of multi-object tracking, the CLEAR 
metrics \cite{clear} have emerged as one of the standards.
By measuring the intersection over union of tracker bounding boxes and matched
ground truth annotations, Accuracy (MOTA) and Precision (MOTP) can be 
computed. Precision measures how well the persons are localized, while 
Accuracy evaluates how many distinct errors such as missed targets (FN), ghost
tracks (FP), or identity switches (IDSW) are made.
Another set of measures that is widely used is 
that of \cite{Li:2009:CVPR}: mostly tracked (MT), mostly lost (ML), and partially
tracked (PT) pedestrians. 
These numbers give a very good intuition on the overall 
performance of the method. 


\subsection{Do Metrics and Humans Agree?}

Our first goal is to find out how well these metrics reflect the perception of a human evaluator on the quality of a tracking result.
To that end, we perform the following study. We create a visual quality assessment task by asking each participant to choose the best tracker among two randomly sampled trackers from the list. The participants only see the video results of the two trackers playing at the same time, and have the possibility to go forward and backward or stop the video at any point. 
In total, we collected results from 500 participants, both from the vision community as well as external visitors of our benchmark website.

\begin{figure}
\centering
\includegraphics[width=.98\linewidth]{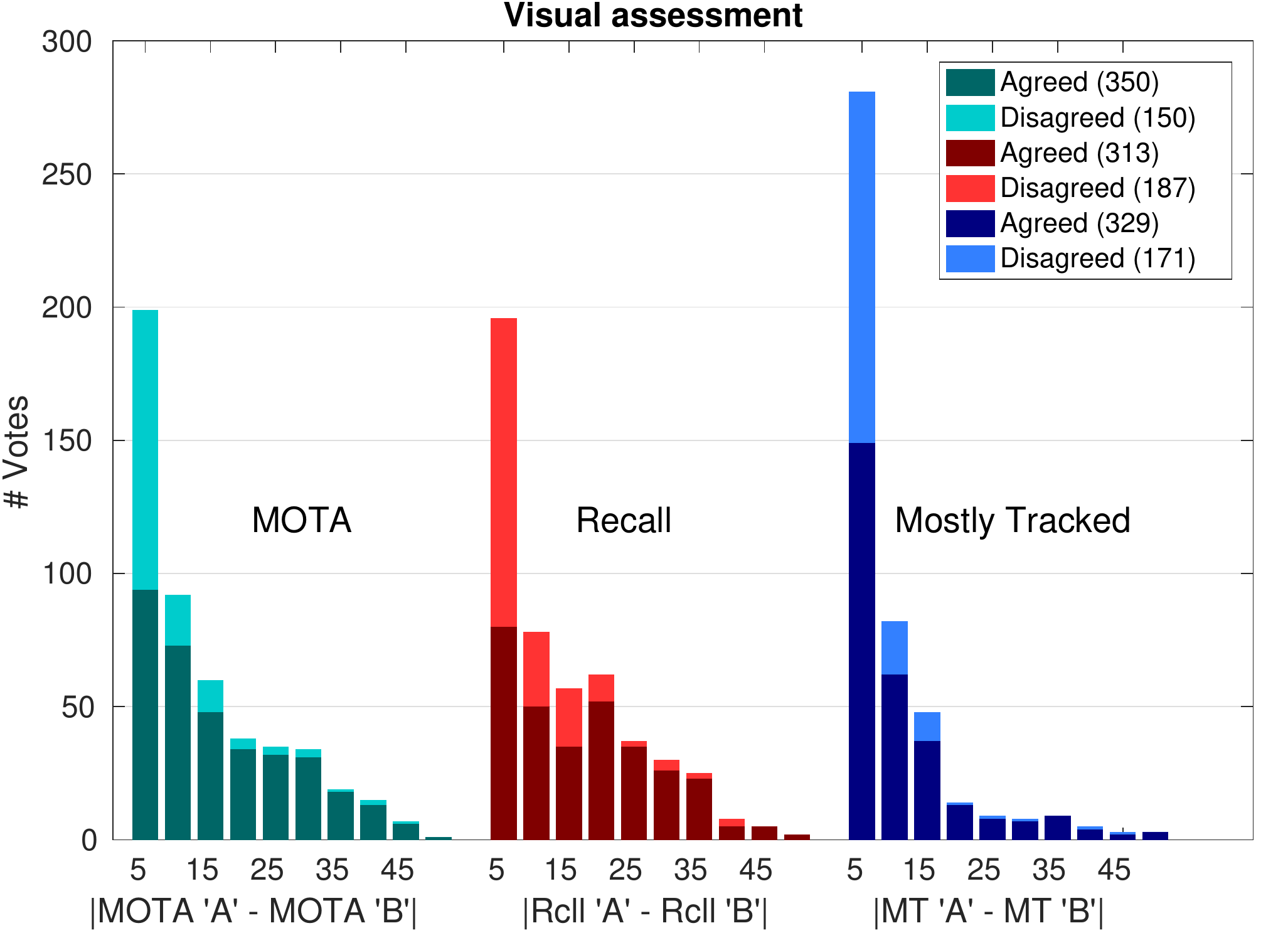}
  \\[1em]
  \includegraphics[width=0.95\linewidth]{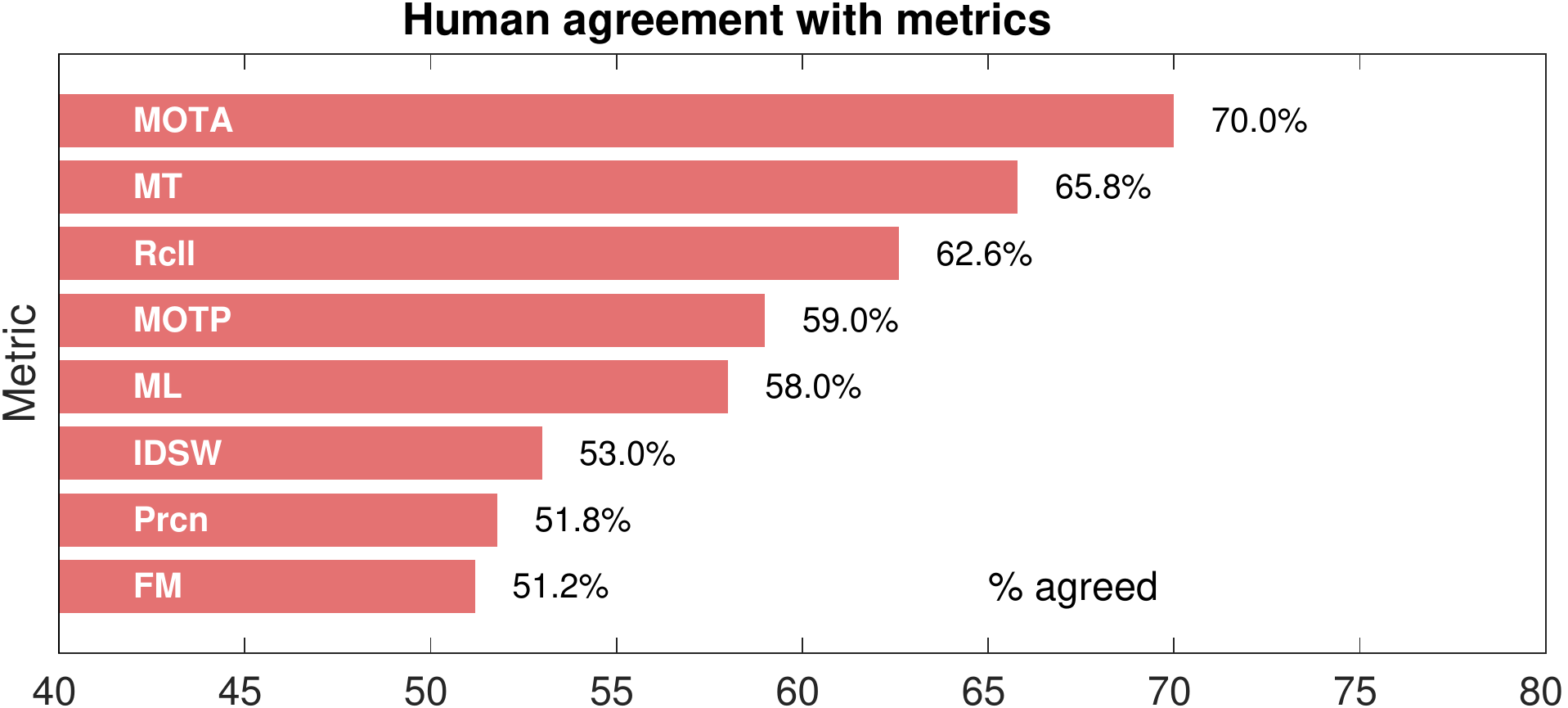}  
  \caption{Results for the tracker visual quality assessment. {\it Top:} Votes from human observers that agree with the metrics (MOTA, Recall and Mostly Tracked) vs the performance difference between tracker A and B under evaluation.   {\it Bottom:} Percentage of votes that align with each of the performance metrics.}
 \label{fig:hon}
 \vspace{-1em}
\end{figure}

Each tracker pair is then ranked according to each measure. In \Fig~\ref{fig:hon} (top), we plot the votes that agreed with that ranking with respect to the difference in the metric. We can see that when the two trackers have a difference of 5 percentage points in MOTA or less, the human visual assessment cannot distinguish which one is better, since the votes are split almost 50/50.
The same observation applies to the Mostly Tracked (MT) metric as well as Recall (Rcll). After that, the opinions quickly align with the metrics and it is easy for a human observer to distinguish the better tracker.
We thus conclude that all trackers with a difference of 5 percentage points or less in MOTA, MT, or Recall can be considered to have a largely equivalent performance.

Another interesting question that we aim to answer with this experiment is which measure reflects the human visual assessment best. In \Fig~\ref{fig:hon} (bottom) we plot the percentage of votes that agree with the metric's assessment. While often highly criticized, MOTA is still the measure that best aligns with the human visual assessment. Mostly Tracked (MT) follows as second-best measure. 
Unsurprisingly, identity switches do not have much of an impact on the quality assessment. This reflects that human observers give much more importance to the fact that people are detected rather than them being correctly tracked.

\subsection{Are All Metrics Necessary?}
\label{sec:metric-importance}
While it is clear that Accuracy (MOTA) and Precision (MOTP) measure two different aspects in tracking, and MOTP is depending mostly on the detector's ability to localize bounding boxes, it is often unclear whether Mostly Tracked (MT) and MOTA actually measure the same characteristics of a tracker. 
In \Fig~\ref{fig:metriccorrelation}, we plot the correlation of several evaluation metrics. Each point represents a result of a tracker in a snippet (\cf~\Sec~\ref{sec:futuresequences}). Each color belongs to one test sequence of \MOTNEW. 
As expected, Recall and MOTA are highly correlated. Recalling the figures from~\Tab\ref{tab:mot16}, we see that the number of missed targets (FN) is typically two to three orders of magnitude higher than FP and ID, which are the other two components of MOTA.
Interestingly, even though both measures are strongly correlated, \Fig~\ref{fig:hon} suggests that humans agree more with the MOTA measure than with the Recall when asked to visually judge a tracker.

MOTA and MT, on the other hand, do not have such a distinct linear correlation. The points are clearly clustered by sequence, which might indicate that one of the measures actually depends on the scene under evaluation. 
The trend in the community is to report both measures, which is also supported by this experiment.
As expected, identity switches and MOTA are hardly correlated at all.

\renewcommand{\thumbwidth}{0.32\linewidth}
\begin{figure}
\centering
  \includegraphics[width=\thumbwidth]{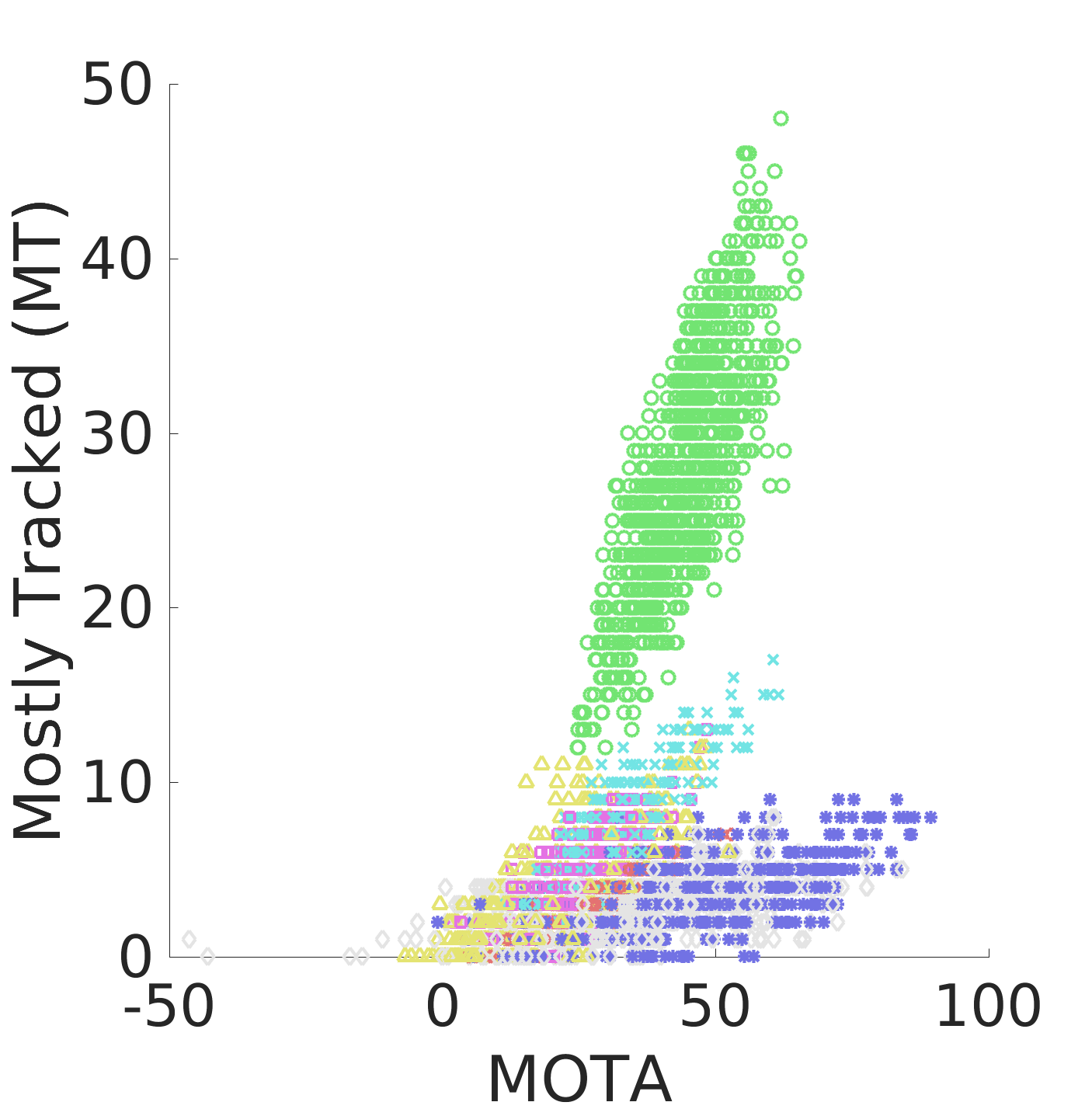}
    \includegraphics[width=\thumbwidth]{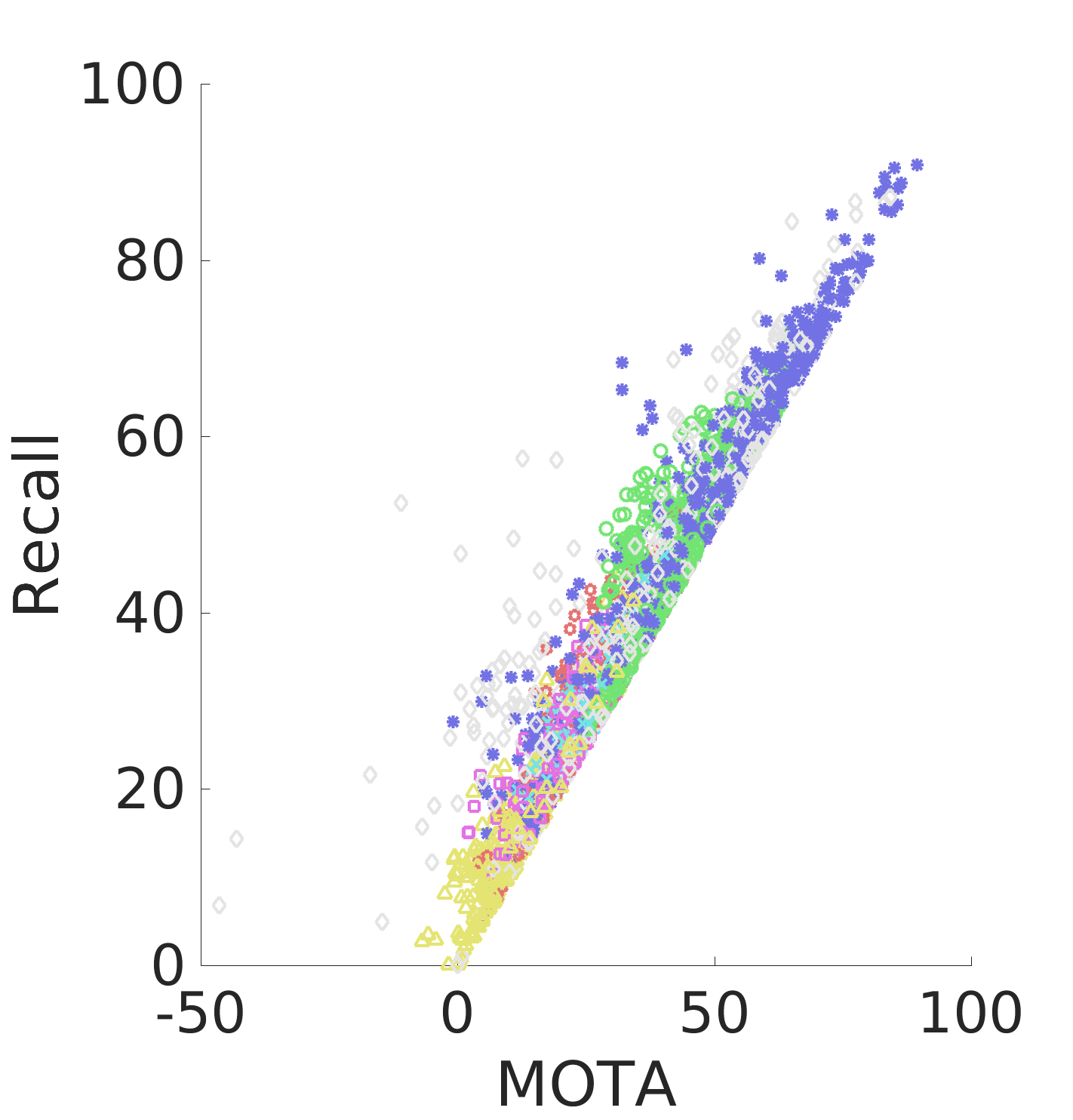}
  \includegraphics[width=\thumbwidth]{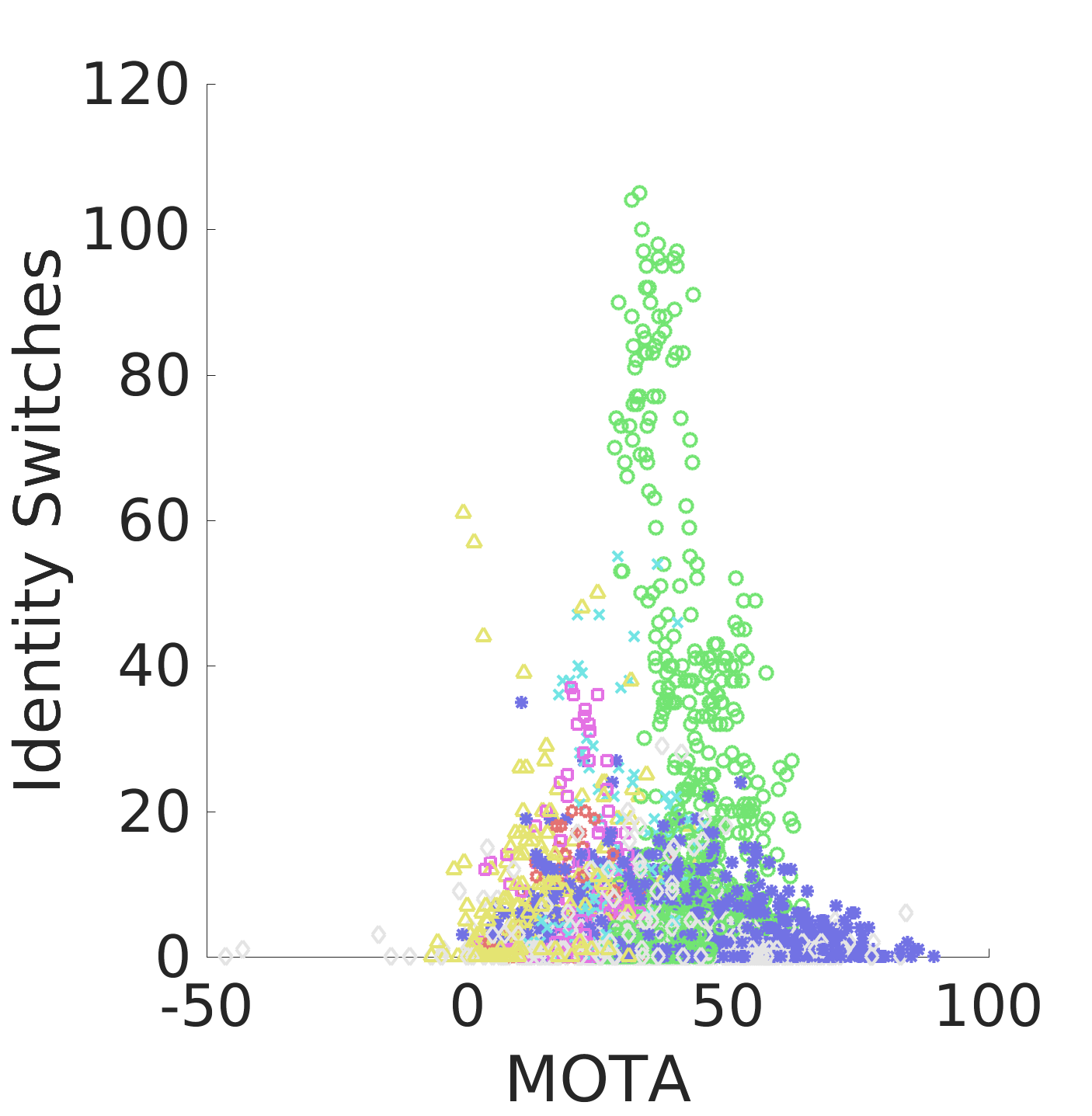}
  \caption{Metric correlations.}
 \label{fig:metriccorrelation}
 \vspace{-1em}
\end{figure}

\subsection{Limitations of the Experiment}
\label{sec:limitations-hon}
Assessing the performance of a multi-object tracking method can vary significantly depending on the application at hand. In surveillance scenarios, it is typically more important to have no false negatives so that no person is missed, while a few false alarms can be easily controlled by humans in the loop.
For applications such as crowd analysis or people motion analysis, it is desirable to have a few very reliable trajectories so that trajectory analysis is accurate, thereby, false positives are not desired. 
In sports, it is crucial to maintain the players' identities robustly to obtain the most reliable statistics.
When creating a benchmark for all-purpose multiple people tracking, it is less clear which metrics to focus on. 
In absence of a concrete application, human judgement is a useful point of reference. 

Of course, the proposed experiment has its limitations. Results in crowded scenes consist of several tens of bounding boxes following numerous pedestrians of different sizes. Studies~\cite{pylyshynspatial1988,alvarezvision2007} have shown that humans are able to track an average of 4 objects at normal speed, reaching up to 8 objects as very slow speeds but are limited to possibly a single object at very high speeds. 
Even though the human observer can stop/restart the video at any time, the complexity of the task remains very high for a human observer. It is therefore likely that he/she will miss small differences between two similarly performing trackers. 

Another issue is regarding the relatively low importance that observers give to identity switches when judging a tracker.
One reason for this could also be due to visualization. Each image is overlaid with the bounding box of the detected person and the color of the bounding box represent its identity. While a false positive can be easily spotted by seeing bounding boxes that are overlaid to background rather than person (similar for false negatives), an identity switch is just shown as a change of color in the bounding box, possibly after a long gap, which might not be perceived by the human observer.
Despite the aforementioned limitations, the experiment offers valuable insights that match the community's perception of existing performance metrics.

 \section{Conclusion and Future Work}
 \label{sec:conclusion}
 
We have introduced a standardized benchmark for fair evaluation of multi-target tracking methods and its two data releases with circa 23000 frames of footage and almost 400000 annotated pedestrians.
We have analyzed the commonly used evaluation metrics with an experiment with human evaluators, and found that even though MOTA have been frequently criticized by the community, it still remains the most representative measure that coincides to the highest degree with human visual assessment.
We have further analyzed the performance of 32 state-of-the-art trackers on \MOTOLD and 16 trackers on \MOTNEW, obtaining several insights. In particular, that trackers performance is very influenced by the affinity metrics used, and deeply learned models are currently giving the most encouraging results. 
Furthermore, we found the expected performance of different approaches highly correlated across videos, \ie most methods perform similarly well or poorly on the same video sequence or fragment.
We believe that our Multiple Object Tracking Benchmark and the presented systematic analysis of existing tracking algorithms helps to identify their strengths and weaknesses and paves the way for future innovations.\\


\noindent{\bf Acknowledgements.} We would like to specially acknowledge Siyu Tang, Sarah Becker, Andreas Lin and Kinga Milan for their help in  the annotation process.
IDR gratefully acknowledges the support of the  Australian Research Council through FL130100102.
LLT and DC were supported by the ERC Consolidator Grant {\em 3D Reloaded}.
SR was supported by the European Research Council under the European Union's Seventh Framework Programme (FP7/2007- 2013) / ERC Grant Agreement No. 307942.

{\small
\bibliographystyle{ieee}
\bibliography{refs-short,refs-new,refs-anton,refs-lau}
}

\clearpage

\begin{appendices}

\twocolumn[\centering\section*{\Large Supplementary Material}]
The following sections contain additional material that was not included in the main paper. In particular,  we first provide a more thorough description of the two datasets and the annotation rules for \MOTNEW. We also include full leaderboards for both releases listing all metrics of the 48 analyzed trackers. 
Finally, we provide additional figures for the error analysis and the fine-grained analysis. \\



\section*{The Multi-Object Tracking Benchmark}
\textbf{\MOTOLD} includes a total of 22 sequences, of which we use half for 
training and half for testing. Most of these sequences had been previously introduced and used by the multi-target tracking community. For \MOTOLD we rely on the publicly available ground truth.
In Table \ref{tab:mot15data}, we detail the characteristics of each of the sequences including length of the sequence, number of pedestrian tracks, number of pedestrian bounding boxes, density of the scene, moving or static camera, viewpoint, and weather conditions.

\begin{table*}[!h]
\small
\begin {center}
 \begin{tabular}{|l| c| c| c| c| c| c| c  | c | c| }
 \hline
 \multicolumn{10}{|c|}{\bf Training sequences} \\ 
 \hline 
      Name & FPS & Resolution & Length & Tracks & Boxes & Density  & Camera & Viewpoint & Conditions \\ 
      \hline
     TUD-Stadtmitte & 25 & 640x480 & 179 (00:07) & 10 & 1156 & 6.5  & static  & medium & normal \\
     TUD-Campus & 25 & 640x480 & 71 (00:03) & 8 & 359 & 5.1 &  static & medium & normal \\
    PETS09-S2L1 & 7 & 768x576 & 795 (01:54) & 19 & 4476 & 5.6 &  static & high & normal \\
    ETH-Bahnhof & 14 & 640x480 & 1000 (01:11) & 171 & 5415 & 5.4 &  moving & low & normal	\\
    ETH-Sunnyday & 14 & 640x480 & 354 (00:25) & 30 & 1858 & 5.2 & moving & low & shadows\\
     ETH-Pedcross2 & 14 & 640x480 & 840 (01:00) & 133 & 6263 & 7.5 & moving & low & shadows \\
     ADL-Rundle-6 & 30 & 1920x1080 & 525 (00:18) & 24 & 5009 & 9.5  & static & low & indoor  \\
           ADL-Rundle-8 & 30 & 1920x1080 & 654 (00:22) & 28 & 6783 & 10.4  & moving & medium & night \\
      KITTI-13 & 10 & 1242x375 & 340 (00:34) & 42 & 762 & 2.2 &moving & medium & shadows \\
       KITTI-17 & 10 & 1242x370 & 145 (00:15) & 9 & 683 & 4.7  & static & medium & shadows \\
      Venice-2 & 30 & 1920x1080 & 600 (00:20) & 26 & 7141 & 11.9 & static & medium & normal \\
      \hline
      \multicolumn{3}{|c|}{\bf Total training} & {\bf 5503 (06:29)} & {\bf 500} & {\bf 39905} & {\bf 7.3}  & &  &  \\
    \hline
    \multicolumn{10}{c}{\vspace{0.25em}} \\
    \hline
     \multicolumn{10}{|c|}{\bf Testing sequences} \\ 
     \hline 
 Name & FPS & Resolution & Length & Tracks & Boxes & Density  &  Camera & Viewpoint & Conditions \\ 
 \hline
     TUD-Crossing & 25 & 640x480 & 201 (00:08) & 13 & 1102 & 5.5 & static  & medium & normal \\  
     PETS09-S2L2 & 7 & 768x576 & 436 (01:02) & 42 & 9641 & 22.1  & static & high & normal \\
     ETH-Jelmoli & 14 & 640x480 & 440 (00:31) & 45 & 2537 & 5.8 & moving & low & shadows	\\
    ETH-Linthescher & 14 & 640x480 & 1194 (01:25) & 197 & 8930 & 7.5  & moving & low & shadows	\\
     ETH-Crossing & 14 & 640x480 & 219 (00:16) & 26 & 1003 & 4.6& moving & low & normal \\
     AVG-TownCentre & 2.5 & 1920x1080 & 450 (03:45) & 226 & 7148 & 15.9  & static & high & normal \\
          ADL-Rundle-1 & 30 & 1920x1080 & 500 (00:17) & 32 & 9306 & 18.6  & moving & medium & normal \\
      ADL-Rundle-3 & 30 & 1920x1080 & 625 (00:21) & 44 & 10166 & 16.3  & static & medium & shadows\\
      KITTI-16 & 10 & 1242x370 & 209 (00:21) & 17 & 1701 & 8.1  & static & medium & shadows\\
       KITTI-19 & 10 & 1242x374 & 1059 (01:46) & 62 & 5343 & 5.0& moving & medium & shadows \\
       Venice-1 & 30 & 1920x1080 & 450 (00:15) & 17 & 4563 & 10.1  & static & medium & normal \\
       \hline
      \multicolumn{3}{|c|}{\bf Total testing} & {\bf 5783 (10:07)} & {\bf 721} & {\bf 61440} & {\bf 10.6} & & &  \\
      \hline 
    \end{tabular}
  \end{center}
    \caption{Overview of the sequences included in the \MOTOLD release.}
\label{tab:mot15data}
\end{table*}


In contrast to the 2015 edition, \textbf{\MOTNEW} consists almost exclusively of novel, high-definition videos, \emph{all} of which have been (re-)annotated following a consistent protocol (see Section~\ref{sec:anno-rules}). 
\MOTNEW includes a total of 14 sequences, of which we use half for 
training and half for testing. 
For these sequences, we also provide further information in Table \ref{tab:mot16data}.
\begin{table*}[tbh]
\small
\begin {center}
 \begin{tabular}{|l| c| c| r| r| r| r| c |c | c |  }
 \hline
 \multicolumn{10}{|c|}{\bf Training sequences} \\ 
 \hline 
      Name & FPS & Resolution & Length & Tracks & Boxes & Density & Camera & Viewpoint & Conditions \\ 
      \hline
     02 & 30 & 1920x1080 & 600 (00:20) & 49 & 17,833 & 29.7  & static  & medium & cloudy  \\
     04 & 30 & 1920x1080 & 1,050 (00:35) & 80 & 47,557 & 45.3  & static & high & night  \\
    05 & 14 & 640x480 & 837 (01:00) & 124 & 6,818 & 8.1 & moving & medium & sunny  \\
    09 & 30 & 1920x1080 & 525 (00:18) & 25 & 5,257 & 10.0  & static & low & indoor	 \\
    10 & 30 & 1920x1080 & 654 (00:22) & 54 & 12,318 & 18.8  & moving & medium & night	 \\
    11 & 30 & 1920x1080 & 900 (00:30) & 67 & 9,174 & 10.2 & moving & medium & indoor \\
    13 & 25 & 1920x1080 & 750 (00:30) & 68 & 11,450 & 15.3  & moving & high & sunny  \\
      \hline
      \multicolumn{3}{|c|}{\bf Total training} & {\bf 5,316 (03:35)} & {\bf 512} & {\bf 110,407} & {\bf 20.8} & & &   \\
    \hline
    \multicolumn{10}{c}{\vspace{0.25em}} \\
    \hline
     \multicolumn{10}{|c|}{\bf Testing sequences} \\ 
     \hline 
 Name & FPS & Resolution & Length & Tracks & Boxes & Density &  Camera & Viewpoint & Conditions \\ 
 \hline
       01 & 30 & 1920x1080 & 450 (00:15) & 23 & 6,395 & 14.2  & static  & medium & cloudy  \\
     03 & 30 & 1920x1080 & 1,500 (00:50) & 148 & 104,556 & 69.7  & static & high & night\\
   06 & 14 & 640x480 & 1,194 (01:25) & 217 & 11,538 & 9.7 & moving & medium & sunny \\
    07 & 30 & 1920x1080 & 500 (00:17) & 55 & 16,322 & 32.6  & moving & medium & shadow	 \\
   08 & 30 & 1920x1080 & 625 (00:21) & 63 & 16,737 & 26.8  & static & medium & sunny	 \\
    12 & 30 & 1920x1080 & 900 (00:30) & 94 & 8,295 & 9.2 & moving & medium & indoor \\
   14 & 25 & 1920x1080 & 750 (00:30) &230 & 18,483 & 24.6  & moving & high & sunny  \\
       \hline
      \multicolumn{3}{|c|}{\bf Total testing} & {\bf 5,919 (04:08)} & {\bf 830} & {\bf 182,326} & {\bf 30.8} & & &  \\
      \hline 
    \end{tabular}
  \end{center}
    \caption{Overview of the sequences currently included in the \MOTNEW benchmark.}
\label{tab:mot16data}
\end{table*}
In Table \ref{tab:dataclasses}, we list the types of annotations that we provide with the dataset, including cars, motorbikes, and bicycles in addition to pedestrians.

\begin{table*}[tbh]
\small
\begin {center}
 \begin{tabular}{|l| R{1.1cm}|R{0.9cm}|R{1cm}|R{0.9cm}|R{0.8cm}|R{0.4cm}|R{0.8cm}|R{0.8cm}|R{0.9cm}|R{0.9cm}|R{0.5cm}|R{1.1cm}|}
 \hline
 \multicolumn{13}{|c|}{\bf Annotation classes} \\ 
 \hline 
\rotatebox{90}{Sequence } & 
\rotatebox{90}{Pedestrian} & 
\rotatebox{90}{Person on vehicle}   &  
\rotatebox{90}{Car} & 
\rotatebox{90}{Bicycle} & 
\rotatebox{90}{Motorbike} & 
\rotatebox{90}{Non-motorized vehicle } & 
\rotatebox{90}{Static person} & 
\rotatebox{90}{Distractor}  & 
\rotatebox{90}{Occluder on  ground } & 
\rotatebox{90}{Occluder full}      & 
\rotatebox{90}{Reflection}   & 
\rotatebox{90}{Total} \\
 \hline
  01 & 6,395 & 346 & 0 & 341 & 0 & 0 & 4,790 & 900  &   3,150  & 0   &  0  & 15,922 \\ 
   02  &  17,833 & 1,549  &  0   &  1,559 &  0  & 0  & 5,271  &  1,200 &  1,781  &  0  &  0  &    29,193\\
  03  & 104,556  &  70  &  1,500  &   12,060  &  1,500     & 0   &  6,000   &  0     &  24,000    & 13,500    & 0   &  163,186\\
  04  &  47,557   &    0   &  1,050   &  11,550  &   1,050  &   0  &  4,798  &   0   &   23,100  &  18,900  &  0  &    108,005\\
  05 & 6,818 & 315   &  196 & 315 &  0   &  11  & 0  &  16  &   0    &   0 &  0  &   7,671\\
  06 & 11,538  &   150  &  0     &  118  &     0    & 0   &   269  & 238 & 109     &  0  &   0      & 12,422\\
  07  & 16,322   &   0   &  0  &  0  &  0    &  0  & 2,023   &  0  & 1,920     &    0    & 0    &  20,265\\
  08 & 16,737  & 0  &  0   & 0 & 0   &  0  & 1,715 &  2,719 &   6,875  & 0  &  0  &     28,046\\
  09  & 5,257  &  0  &  0     & 0  &  0 &  0 &  0   & 1,575 &  1,050  &   0   &  948     & 8,830\\
  10  & 12,318  &  0  &  25  &  0    &  0 &  0  &   1,376 &  470   &   2,740   &  0  & 0    &   16,929\\
  11  &  9,174   & 0  &  0 &  0   &  0  & 0  &  0   &  306  &   596  & 0  & 0      & 10,076\\
  12  &   8,295  &  0  &  0   & 0   & 0    &  0 & 1,012  &  765 &    1,394    &  0     &  0   &   11,464\\
  13   &  11,450   &    0  & 4,484   & 103   &  0    &  0    &  0   &  4    &   2,542  &    680     &    0   & 19,263\\
  14  &  18,483    &   0   &  1,563    &     0  &    0 &   0   & 712  &   47   &  4,062     &     393   &  0   &     25,260\\
\hline
  Total & 292,733  & 2,430  & 8,818 & 26,046 &  2,550    & 11  &  27,966 & 8,238   & 73,319  & 33,473  & 948  &   476,532\\
      \hline 
    \end{tabular}
  \end{center}
    \caption{Overview of the types of annotations currently found in the \MOTNEW benchmark.}
\label{tab:dataclasses}

\end{table*}


\section*{Annotation Rules}
\label{sec:anno-rules}

To mitigate the effect of poor and inconsistent labeling, for \MOTNEW, we follow a set of rules to annotate every moving person or vehicle within each sequence 
with a bounding box as accurately as possible. In the following we 
define a clear protocol that was obeyed throughout the entire dataset to 
guarantee consistency.

\subsection*{Target class}
In this benchmark we are interested in tracking moving objects in videos. In particular, we are interested in evaluating multiple people tracking algorithms, therefore, people will be the center of attention of our annotations. 
We divide the pertinent classes into three categories:
\begin{enumerate}
\item {\it moving} or {\it standing} pedestrians;
 \item people that are {\it not in an upright position} or artificial representations of humans;  and 
 \item {\it vehicles} and {\it occluders}.
 \end{enumerate}
 
In the first group, we annotate all moving or standing (upright) pedestrians that appear in the field of view and can be determined as such by the viewer. People on bikes or skateboards will also be annotated in this category (and are typically found by modern pedestrian detectors). Furthermore, if a person \emph{briefly} bends over or squats, \eg to pick something up or to talk to a child, they shall remain in the standard \emph{pedestrian} class.
The algorithms that submit to our benchmark are expected to track these targets.

In the second group we include all people-like objects whose exact classification is ambiguous and can vary depending on the viewer, the application at hand, or other factors. We annotate all static people that are not in an upright position, \eg sitting, lying down. We also include in this category any artificial representation of a human that might cause a strong detection response, such as mannequins, pictures, or reflections. People behind glass should also be marked as distractors.
The idea is to use these annotations in the evaluation such that an algorithm is neither penalized nor rewarded for tracking, \eg, a sitting person or a reflection.

In the third group, we annotate all moving vehicles such as cars, bicycles, motorbikes, and non-motorized vehicles (\eg, strollers), as well as other potential occluders. We will not evaluate specifically against these annotations, rather they are provided to the users both for training purposes and for computing the level of occlusion of pedestrians. Static vehicles (parked cars, bicycles) are not annotated as long as they do not occlude any pedestrians.

%
%

\begin{table}[t]
\begin{tabular}{lp{.75\linewidth}}
& Instruction\\
\hline
What? & Targets: All upright people including\\
& + walking, standing, running pedestrians\\
& + cyclists, skaters\\ [1em]
& Distractors: Static people or representations\\
& + people not in upright position (sitting, lying down)\\
& + reflections, drawings, or photographs of people\\
& + human-like objects like dolls, mannequins\\[1em]
& Others: Moving vehicles and occluders\\
& + Cars, bikes, motorbikes\\
& + Pillars, trees, buildings\\
\hline
When? & Start as early as possible.\\
& End as late as possible.\\
& Keep ID as long as the person is inside the field of view and its path can be determined unambiguously.\\
\hline
How? & The bounding box should contain all pixels belonging to that person and at the same time be as tight as possible.\\
\hline
Occlusions & Always annotate during occlusions if the position can be determined unambiguously. \\
& If the occlusion is very long and it is not possible to determine the path of the object using simple reasoning (\eg, constant velocity assumption), the object will be assigned a new ID once it reappears. \\
\hline
\end{tabular}
\smallskip
\caption{Instructions obeyed during annotations.}
\label{tab:instructions}
\end{table}

\begin{figure*}
\centering
 \includegraphics[width=0.6\linewidth]{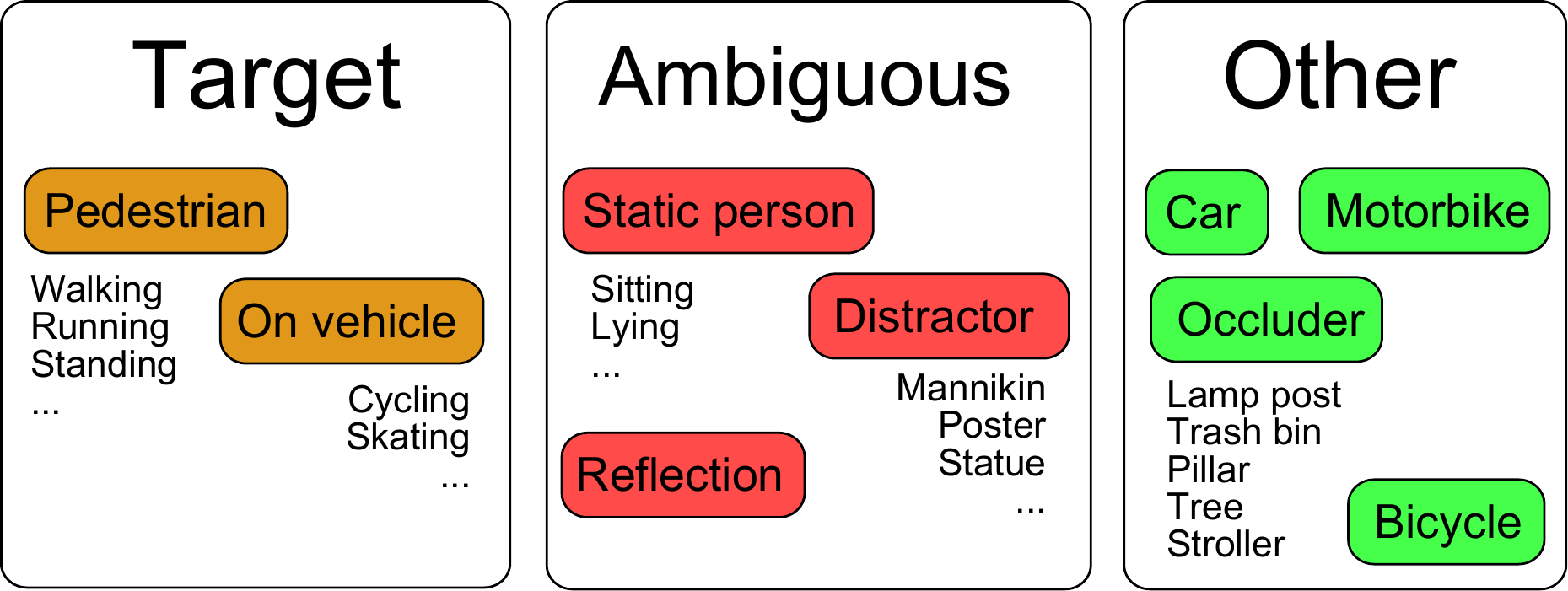}
  \includegraphics[width=0.38\linewidth]{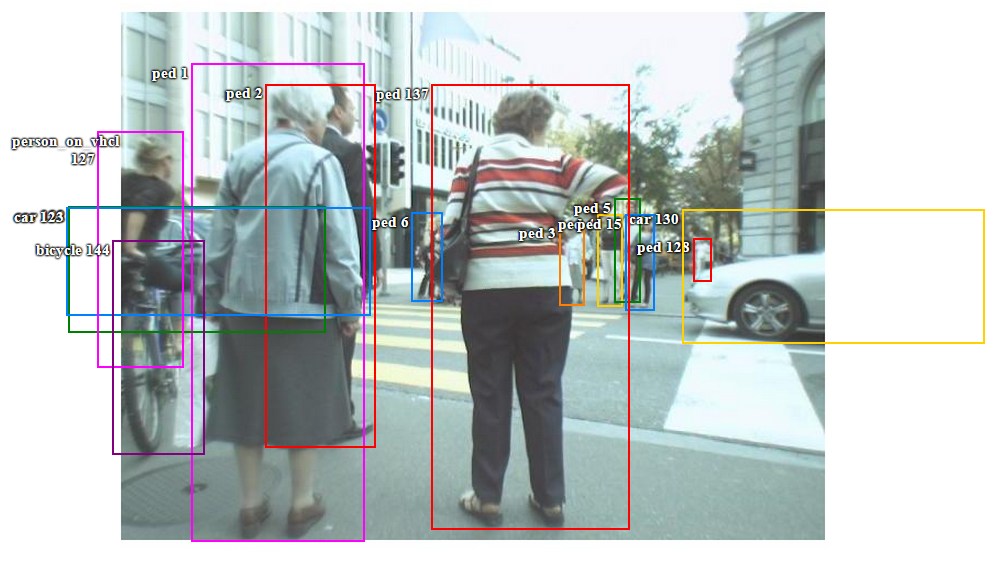}
 \caption{Left: An overview of annotated classes. The classes in orange will be the central ones to evaluate on. The red classes include ambiguous cases such that both recovering and missing them will not be penalized in the evaluation. The classes in green are annotated for training purposes and for computing the occlusion level of all pedestrians. Right: An exemplar of an annotated frame. Note how partially cropped objects are also marked outside of the frame. Also note that the bounding box encloses the entire person but not, \eg, the white bag of Pedestrian 1 (bottom left).}
 \label{fig:class}
\end{figure*}

The rules are summarized in \Tab~\ref{tab:instructions} and in \Fig~\ref{fig:class} we present a diagram of the classes of objects we annotate, as well as a sample frame with annotations. 

\subsection*{Bounding box alignment}
The bounding box is aligned with the object's extent as accurately as possible. 
The bounding box should contain all pixels belonging to that object and at the same time be as tight as possible, \ie no pixels should be left outside the box. 
This means that a walking side-view pedestrian will typically have a box whose width varies periodically with the stride, while a front view or a standing person will maintain a more constant aspect ratio over time. If the person is partially occluded, the extent is estimated 
based on other available information such as expected size, shadows, 
reflections, previous and future frames, and other cues. If a person is 
cropped by the image border, the box is estimated beyond the original frame to represent the entire person and to estimate the level of cropping. If an occluding object cannot be accurately enclosed in one box (\eg, a tree with branches or an escalator may require a large bounding box where most of the area does not belong to the actual object), then several boxes may be used to better approximate the extent of that object.

Persons on vehicles will only be annotated separately from the vehicle if clearly visible. For example, children inside strollers or people inside cars will not be annotated, while motorcyclists or bikers will be.

\subsection*{Start and end of trajectories}
The box (track) appears as soon as the person's location and extent can 
be determined precisely. This is typically the case when $\approx 10 \%$ of the person becomes visible.
  Similarly, the track ends when it is no longer 
possible to pinpoint the exact location. In other words the annotation 
starts as early and ends as late as possible such that the accuracy is 
not forfeited. The box coordinates may exceed the visible area.
Should a person leave the field of view and appear at a 
later point, they will be assigned a new ID.

\subsection*{Minimal size}
Although the evaluation will only take into account pedestrians that have a minimum height in pixels, 
annotations will contain all objects of all sizes as long as they are distinguishable by the annotator. In other words, \emph{all} targets independent of their size on the image shall be annotated.

\subsection*{Occlusions}
There is no need to explicitly annotate the level of occlusion. This value will be computed automatically using the ground plane assumption and the annotations. Each target is fully annotated through occlusions as long as its extent and location can be determined accurately enough. If a target becomes completely occluded in the middle of the sequence and does not become visible later, the track should be terminated (marked as `outside of view'). If a target reappears after a prolonged period such that its location is ambiguous during the occlusion, it will reappear with a new ID.

\subsection*{Sanity check}
Upon annotating all sequences, a ``sanity check'' was carried out to ensure that no relevant entities were missed. To that end, we ran a pedestrian detector on all videos and added all high-confidence detections that corresponded to either humans or distractors to the annotation list.

\section*{Runtime Analysis}
\label{sec:runtime}

Different methods require a varying degree of computational resources to perform the task of multi-target tracking. In our current setup, we only consider the final tracking output in form of labeled bounding boxes and therefore cannot directly measure the efficiency of a particular method. Moreover, the efficiency is extremely hard to compare because some methods may require large amounts of memory, others prefer a multi-core system, while others still can be easily executed on a GPU.  For our purpose, we ask each benchmark participant to provide the number of seconds required to produce the results on the entire dataset. The resulting numbers are therefore only indicative of each approach and are not immediately comparable to one another.

\begin{figure*}[thb]
\centering
  \includegraphics[width=.47\linewidth]{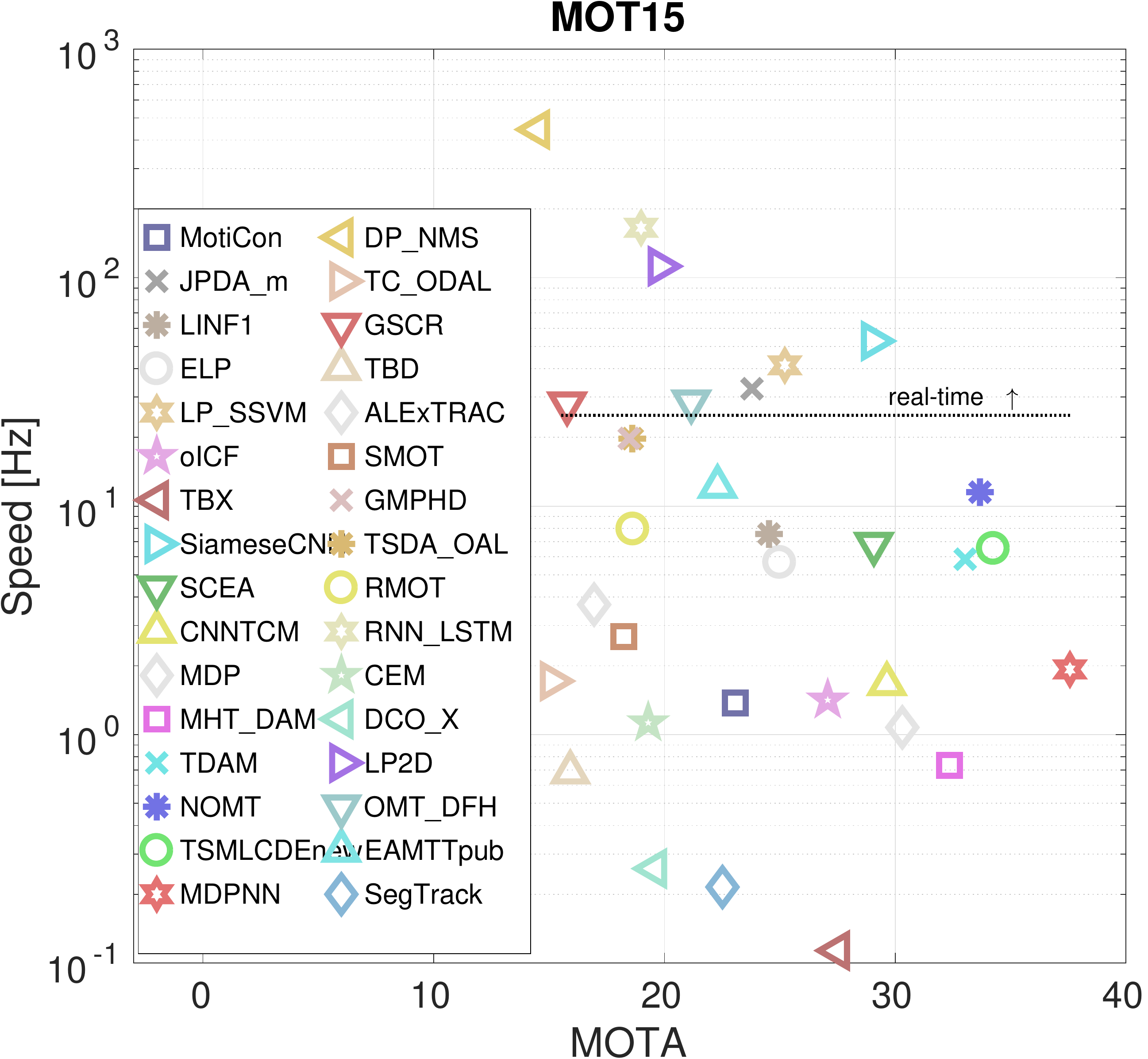}
  \hfill
  \includegraphics[width=.48\linewidth]{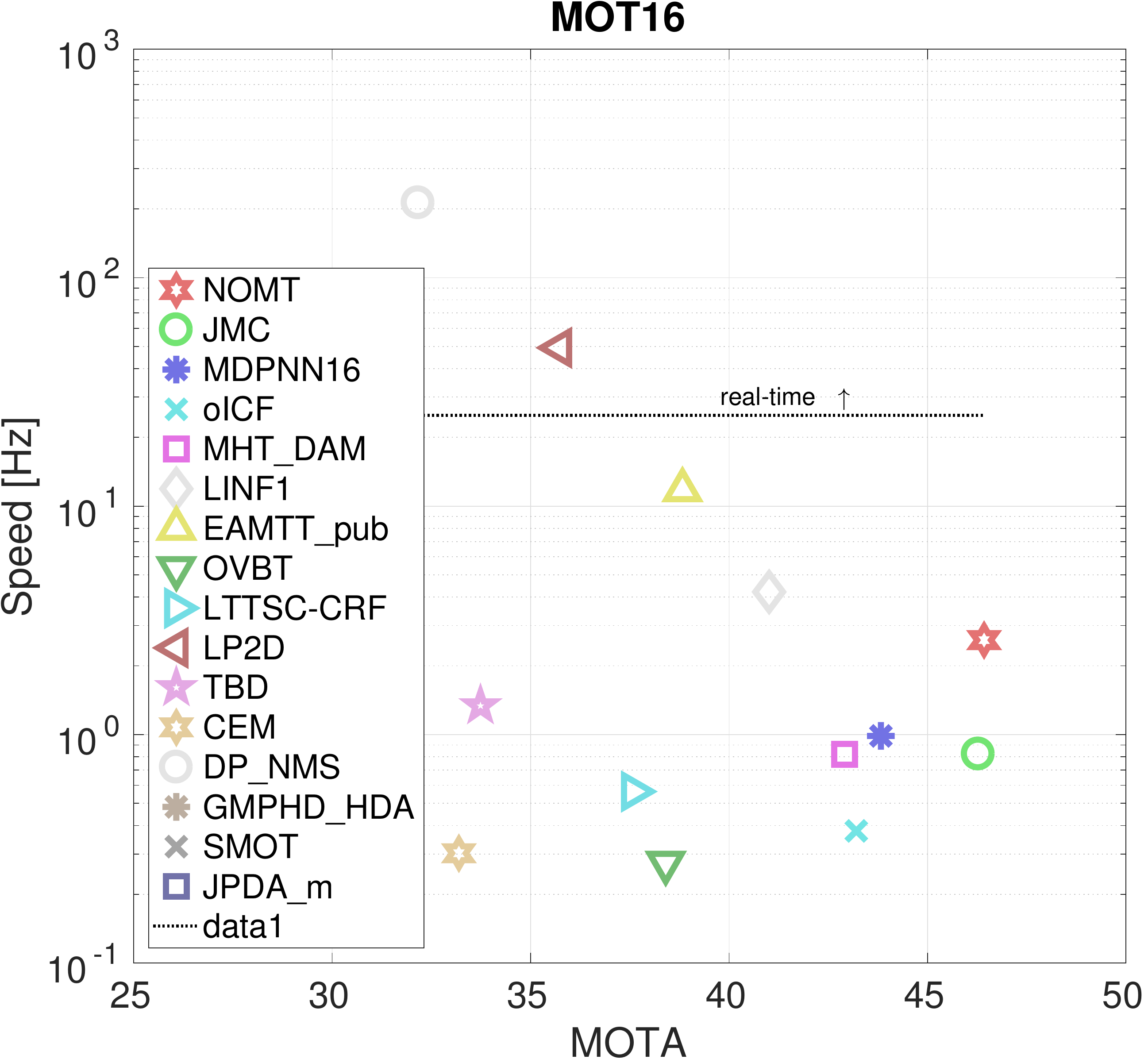}
  \caption{Tracker performance measured by MOTA \vs processing efficiency in frames per second for \MOTOLD (left) and \MOTNEW on a log-scale. The latter is only indicative of the true value and has not been measured by the benchmark organizers. See text for details.}
 \label{fig:MOTA-runtime}
\end{figure*}

\Fig~\ref{fig:MOTA-runtime} shows the relationship between each submission's performance measured by MOTA and its efficiency in terms of frames per second, averaged over the entire dataset. There are two observations worth pointing out. First, with very few exceptions, the majority of methods is still below real-time performance, which is assumed at 25 Hz. Second, the average processing is slower for the \MOTNEW dataset, which is most likely due to its higher average density of 26 persons per frame, compared to only 9 in \MOTOLD. 


\section*{Error Analysis}
\label{sec:supp-error-analysis}
Here, we provide additional figures for Sections 4 and 5 of the main paper.

\subsection*{False negatives \vs false positives}
\label{sec:fn-fp}
\begin{figure}[b]
\centering
  \includegraphics[width=1\linewidth]{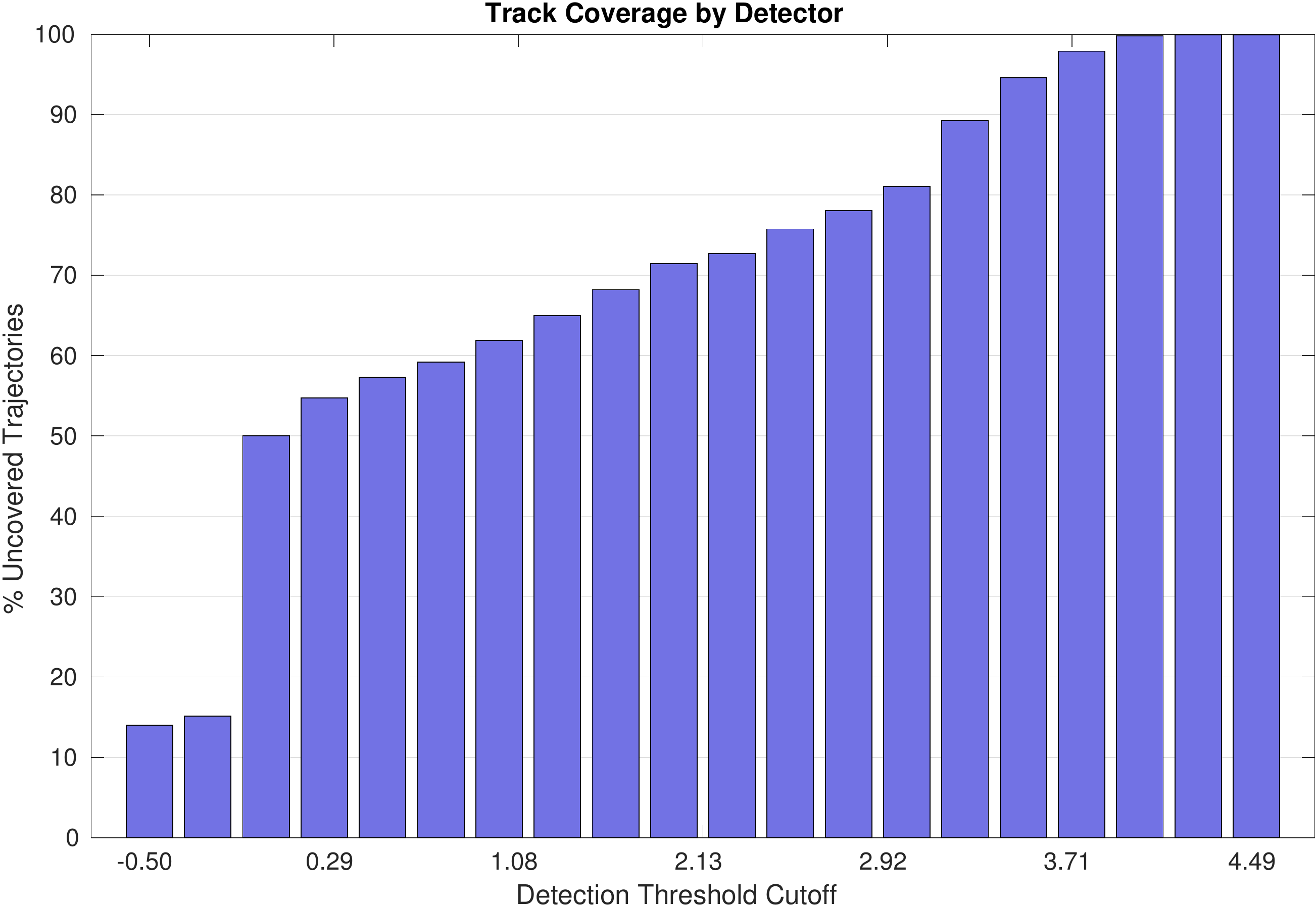}
  \caption{Percentage of covered trajectories (with at least one detection with 50\% overlap or more) when varying the detection threshold cutoff. Note that if we drop the 10\% detections with lowest confidence, more than 50\% of the trajectories are not covered by any detection.}
 \label{fig:det-threshold}
\end{figure}

One of the rather surprising findings of our work is the fact that most tracking methods aim to reduce false alarms produced by the person detector and hardly focus on increasing recall, \ie reducing the number of false negatives. This is evidenced by Figures 2 and 4 in the paper.
\Fig~\ref{fig:det-threshold} visualizes the detector coverage as described in Sec.~4.2. For each detection threshold, we computed the percentage of trajectories covered by detections. A ground truth track is considered `covered' if there is at least one detection with an overlap over $50\%$. Taking all detections with the confidence value above $-0.5$ into account, as many as 18\% of the trajectories are not covered by any detection. What is even more surprising is that if we drop the 10\% detections with the lowest confidence and only consider those with the confidence value above $0.2$, the number of completely uncovered tracks jumps up to 55\%. Therefore we can conclude that reducing the number of false alarms by choosing a more conservative threshold will significantly hurt the tracking performance.


\section*{Fine-grained Analysis}
We include additional plots depicting the fine-grained analysis in \Fig~\ref{fig:snippets}.
\begin{figure*}[t]
\centering
  \includegraphics[width=.48\linewidth]{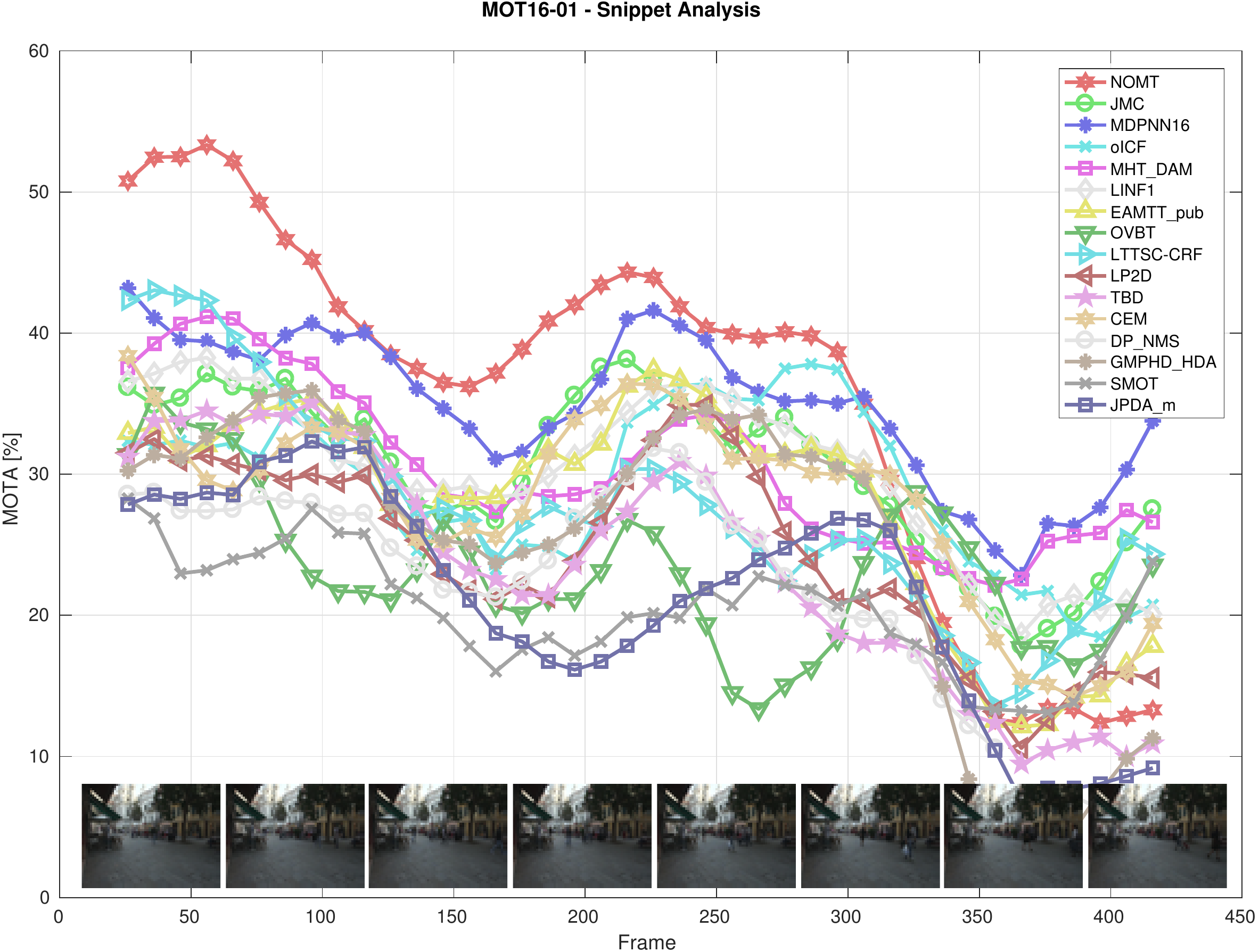}
  \hfill
  \includegraphics[width=.48\linewidth]{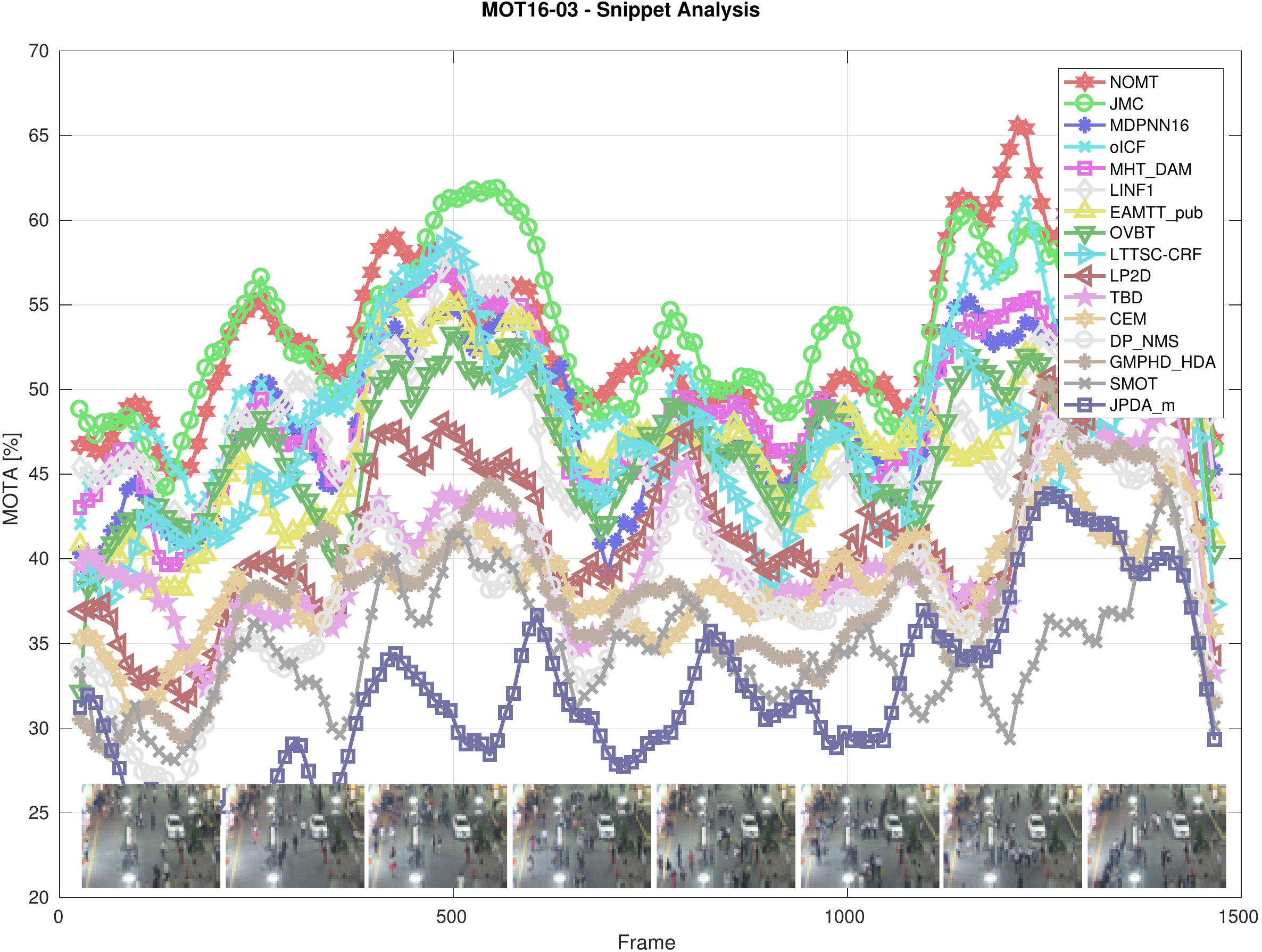}\\  
  \includegraphics[width=.48\linewidth]{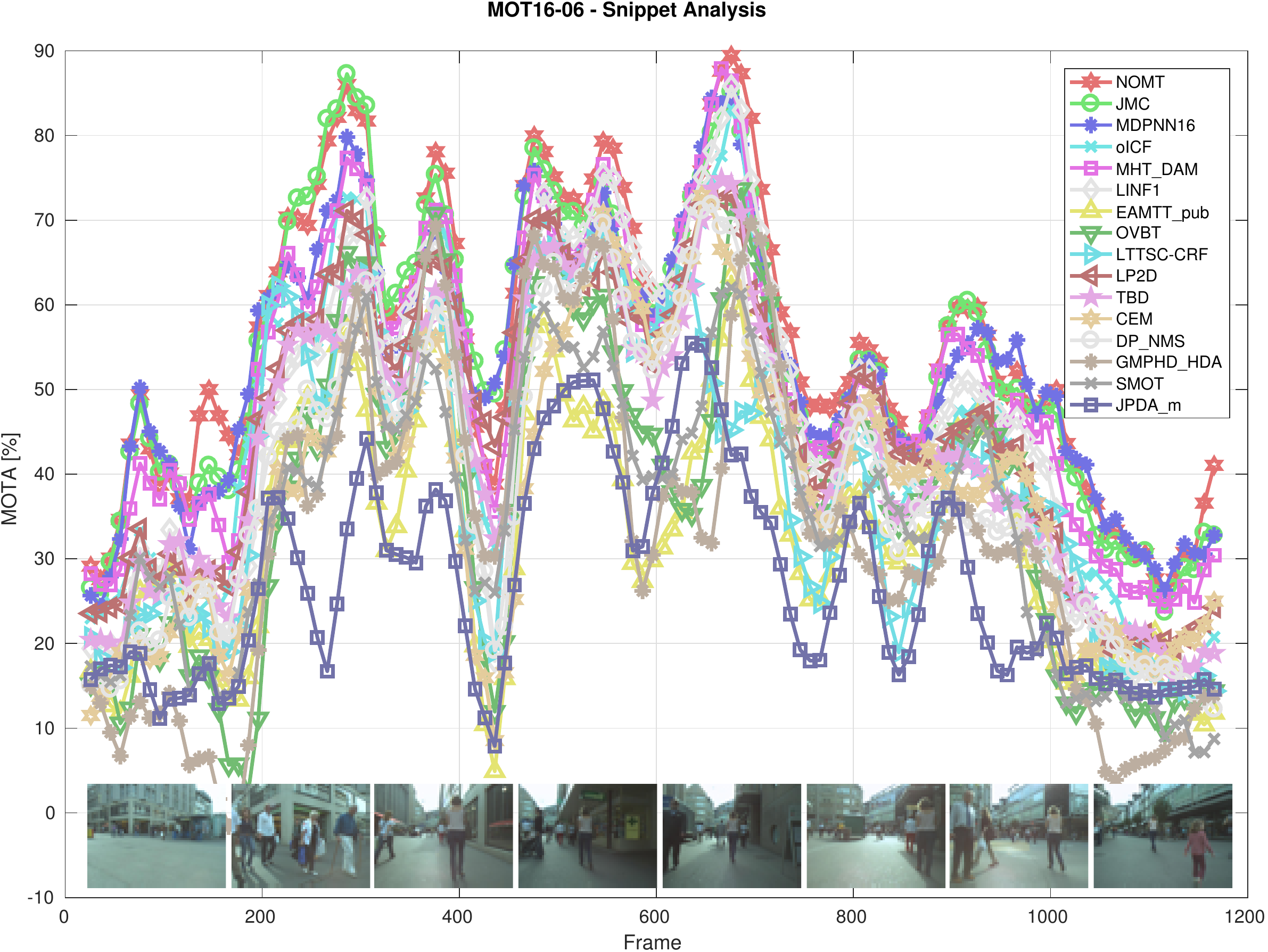}
  \hfill
  \includegraphics[width=.48\linewidth]{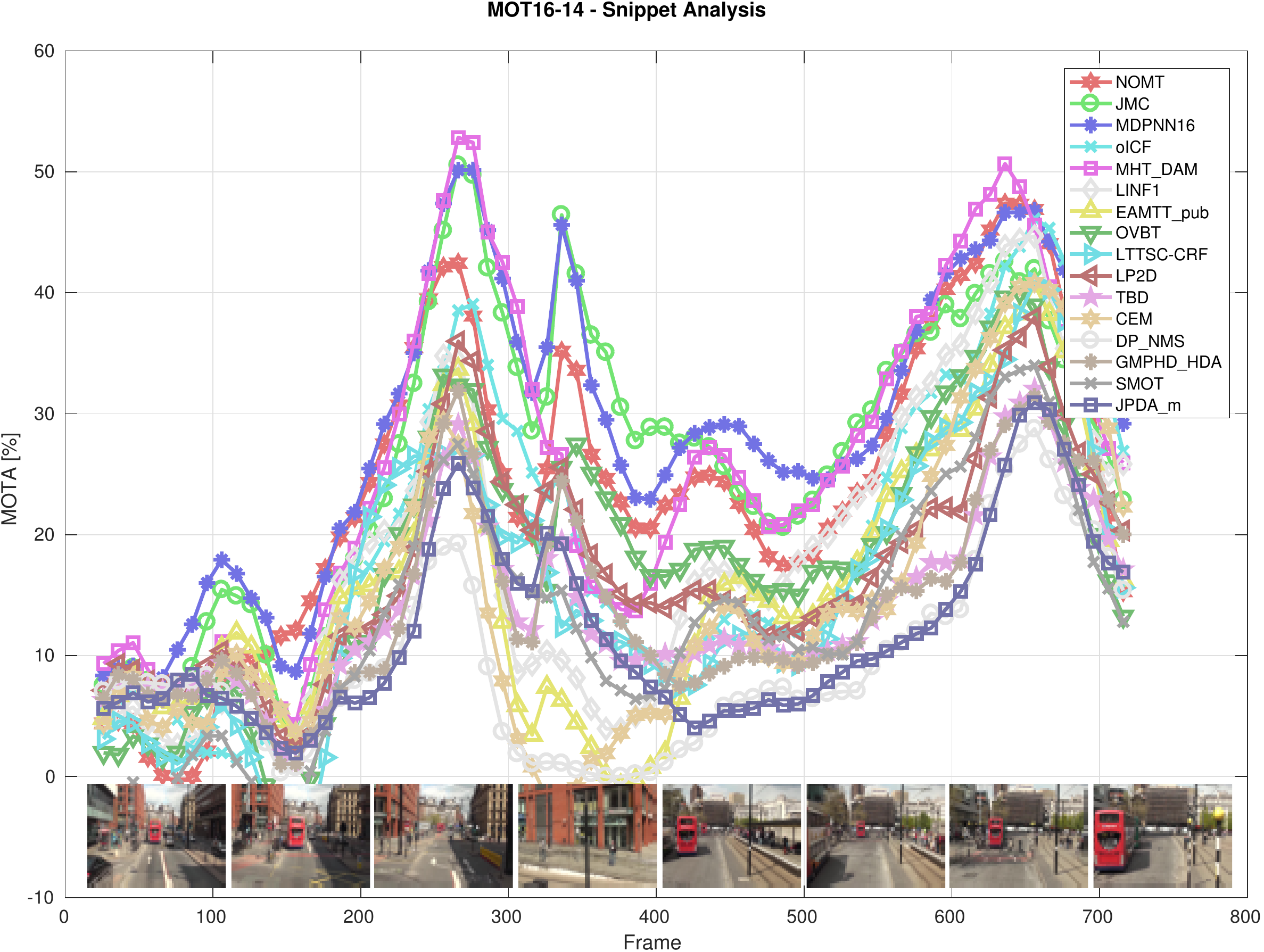}
\caption{Additional examples of fine-grained analysis on four sequences. Each line represents the \emph{local} performance of a tracker, measured by MOTA within a 50-frame (roughly 3 seconds long) segment. Note the extreme within-sequence variation, in particular for the two moving-camera sequences on the bottom.}
 \label{fig:snippets}
\end{figure*}


%
%
%
%
%

\section*{State of the Art in Multi-Object Tracking}
\label{sec:leaderboard}
Detailed tracking results for both \MOTOLD and \MOTNEW datasets are presented in Tables~\ref{tab:mot15} and~\ref{tab:mot16}, respectively. In addition to the standard metrics described in Section 6 of the paper, we also list the false alarm per frame (FAF) rate, and the number of track fragmentations (Frag). A track fragmentation is counted when a ground truth trajectory is lost for any number of frames after successfully being tracked and before tracking is resumed. For MOTA, we also provide the standard deviation across all sequences as an indicator for the stability of the tracker. Next to the raw numbers of ID switches and fragmentations, we also list the relative errors in parentheses, which are computed as IDSw/Recall and Frag/Recall, respectively. This is important to take into account the number of correctly tracked targets.


\begin{table*}[htb]
  \caption{The \MOTOLD leaderboard. Performance of several trackers according to different metrics.}
  \label{tab:mot15}
  \smallskip
\centering
\begin{tabular*}{\linewidth}{@{\extracolsep{\stretch{1}}}l rrrrrrrrr @{}}
\toprule
  Method & MOTA & MOTP & FAF & MT & ML & FP & FN & IDsw & Frag \\ \midrule
 MDPNN \cite{SadeghianArxiv2017} & 36.6 $\pm$ 12.1	& 71.4	& 1.1 &	13.3 &	36.5 &	6419	 & 31811 &	700 (14.5)	& 1458 (30.2) \\
 
 TSMLCDEnew \cite{WangArxiv2015} & 34.3$\pm$13.1	&71.7	&1.4	&14.0&	39.4&	7869	&31908	&618 (12.9)	&959 (20.0) \\

 NOMT \cite{Choi:2015:ICCV} & 33.7$\pm$  16.2	&71.9&	1.3	&12.2&	44.0&	7762&	32547&	442 (9.4)&	823 (17.5) \\
 
 TDAM \cite{MinArxiv2015} 
& 33.0$\pm$9.8 
& 	72.8
& 	1.7	
& 13.3
& 	39.1
& 	10064	
& 30617
& 	464 (9.2)
& 	1506 (30.0) \\
 
 MHT\_DAM \cite{Kim:2015:ICCV} & 32.4$\pm$15.6	&71.8&	1.6	&16.0&	43.8&	9064&	32060&	435 (9.1)	&826 (17.3) \\
 
 MDP \cite{Xiang:2015:ICCV} & 30.3$\pm$ 14.6	& 71.3&	1.7&	13.0&	38.4&	9717	&32422	&680 (14.4)	&1500 (31.8) \\
 
 CNNTCM \cite{WangCVPRW2016} & 29.6$\pm$13.9& 	71.8	& 1.3& 	11.2& 	44.0& 	7786& 	34733& 	712 (16.4)& 	943 (21.7) \\
 
 SCEA \cite{YoonCVPR2016} 
&  29.1$\pm$12.2 
& 	71.1 
& 	1	
& 8.9 
& 	47.3 
& 	6060 
& 	36912	
& 604 (15.1) 
& 	1182 (29.6) \\
 
 SiameseCNN \cite{lealCVPRW2016} & 29.0$\pm$15.1	& 71.2	& 0.9	& 8.5& 	48.4& 	5160& 	37798	& 639 (16.6)& 	1316 (34.2)\\
 
 TbX \cite{henschelArxiv2016} & 27.5$\pm$13.3 &	70.6&	1.4&	10.4&	45.8&	7968	&35810	&759 (18.2)&	1528 (36.6)\\
 
 olCF \cite{kieritzAVSS2016} & 27.1$\pm$14.9& 	70	& 1.3	& 6.4& 	48.7& 	7594	& 36757& 	454 (11.3)& 	1660 (41.3) \\
 
 LP\_SSVM \cite{WangIJCV2016} & 25.2$\pm$13.7	&71.7	&1.4	&5.8&	53.0&	8369	&36932	&646 (16.2)&	849 (21.3)\\
 
  ELP \cite{mclaughlinWACV2015} & 25.0$\pm$10.8	&71.2&	1.3	&7.5&	43.8&	7345&	37344&	1396 (35.6)	&1804 (46.0)\\
 
LINF1 \cite{FagotECCV2016} & 24.5$\pm$15.4	& 71.3	& 1& 	5.5& 	64.6& 	5864& 	40207& 	298 (8.6)	& 744 (21.5)\\
 
  JPDA\_m \cite{Rezatofighi:2015:ICCV} & 23.8$\pm$15.1	&68.2	&1.1	&5.0&	58.1&	6373&	40084&	365 (10.5)	&869 (25.0)\\
 
  MotiCon \cite{lealcvpr2014} & 23.1$\pm$16.4	&70.9	&1.8&	4.7&	52.0&	10404&	35844&	1018 (24.4)&	1061 (25.5)\\
 
    SegTrack \cite{Milan:2015:CVPR} & 22.5$\pm$15.2&	71.7	&1.4&	5.8&	63.9&	7890&	39020&	697 (19.1)&737 (20.2)\\
 
      EAMTTpub \cite{sanchezbmtt2016} & 22.3$\pm$14.2	&70.8&	1.4	&5.4&	52.7&	7924&	38982	&833 (22.8)&	1485 (40.6)\\
 
            OMT\_DFH \cite{JuIJOSAA2017} & 21.2$\pm$17.2	&69.9&	2.3	&7.1&	46.5&	13218	&34657&	563 (12.9)	&1255 (28.8)\\
 
            LP2D \cite{lealiccv2011} & 19.8$\pm$14.2&	71.2&	2&	6.7&	41.2&	11580	&36045&	1649 (39.9)&	1712 (41.4)\\
 
            DCO\_X \cite{milanTPAMI2016} & 19.6$\pm$14.1	& 71.4&	1.8	&5.1&	54.9&	10652&	38232&	521 (13.8)	&819 (21.7)\\
 
            CEM \cite{Milan:2014:PAMI} & 19.3$\pm$17.5 & 	70.7	& 2.5&	8.5&	46.5&	14180	&34591&	813 (18.6)	&1023 (23.4)\\
 
            RMM\_LSTM \cite{milanAAAI2017} & 19.0$\pm$15.2	&71	&2&	5.5&	45.6&	11578&	36706	&1490 (37.0)&	2081 (51.7)\\
 
   RMOT \cite{Yoon:2015:WACV} & 18.6$\pm$17.5&	69.6&	2.2&	5.3&	53.3&	12473	&36835	&684 (17.1)&	1282 (32.0)\\
 
   TSDA\_OAL \cite{KoIET2017} & 18.6$\pm$17.6	&69.7&	2.8&	9.4&	42.3&	16350&	32853&	806 (17.3)&	1544 (33.2)\\
 
   GMPHD\_15 \cite{SongICCE2016} & 18.5$\pm$12.7&	70.9&	1.4	&3.9&	55.3&	7864&	41766&	459 (14.3)&	1,266 (39.5)\\
 
   SMOT \cite{Dicle:2013:ICCV} & 18.2$\pm$10.3	& 71.2& 	1.5	& 2.8& 	54.8& 	8780& 	40310	& 1148 (33.4)& 	2132 (62.0) \\
 
   ALExTRAC \cite{BewleyICRA2016} & 17.0$\pm$12.1	&71.2	&1.6&	3.9&	52.4&	9233&	39933&	1859 (53.1)&	1872 (53.5)\\
 
   TBD \cite {Geiger:2014:PAMI} & 15.9$\pm$17.6	&70.9&	2.6	&6.4&	47.9&	14943&	34777&	1939 (44.7)&	1963 (45.2)\\
 
   GSCR \cite{FagotICIP2015} & 15.8$\pm$10.5&	69.4	&1.3	&1.8&	61.0&	7597&	43633&	514 (17.7)&	1010 (34.8)\\
 
   TC\_ODAL \cite{Bae:2014:CVPR} & 15.1$\pm$15.0& 	70.5	& 2.2& 	3.2& 	55.8& 	12970& 	38538	& 637 (17.1)& 	1716 (46.0)\\
 
   DP\_NMS \cite{Pirsiavash:2011:CVPR} & 14.5$\pm$13.9&	70.8&	2.3	&6.0&	40.8&	13171&	34814&	4537 (104.7)&	3090 (71.3)\\

\bottomrule
\end{tabular*}
\end{table*}


\end{appendices}

\end{document}